%% file: LEO_iclr2024_conference.tex
\documentclass{article} 
\usepackage{iclr2024_conference,times}

\input{00_imports}

\input{01_front_matter_ICLR}

\begin{document}

\maketitle

\input{02_abstract}

\section{Introduction}
\input{1_introduction}

\section{Methodology}
\input{2_methodology}

\section{Numerical Results}
\input{3_numerical_results}

\section{Discussion}
\input{4_discussion}

\section{Conclusions}
\input{5_conclusion}

\section*{Acknowledgements}
\input{6_acknowledgement}

\bibliography{LEO_iclr2024_conference}
\bibliographystyle{iclr2024_conference}

\clearpage
\appendix
\section{Appendix: Problem formulation}
\input{7_appendix}

\clearpage
\section{Appendix: Prompts}
\input{8_promptappendix}

\end{document}

%% file: 00_imports.tex
\usepackage{lineno,hyperref}
\usepackage{amssymb}
\usepackage{amsmath}
\usepackage{amsthm}
\usepackage{geometry} 
\usepackage{array}    
\usepackage{mathrsfs}
\usepackage{graphicx}
\usepackage{subcaption}
\usepackage{array}
\usepackage[T1]{fontenc}

\usepackage{algorithm2e,setspace}

\usepackage{soul}

\usepackage{booktabs}

\usepackage{tabularx}
\usepackage{multirow}
\usepackage{adjustbox}

\modulolinenumbers[5]

\input{math_commands.tex}

\usepackage{hyperref}
\usepackage{url}

%% file: math_commands.tex
\usepackage{amsmath,amsfonts,bm}

\def\eqref#1{equation~\ref{#1}}

\def\1{\bm{1}}

\DeclareMathAlphabet{\mathsfit}{\encodingdefault}{\sfdefault}{m}{sl}
\SetMathAlphabet{\mathsfit}{bold}{\encodingdefault}{\sfdefault}{bx}{n}



%% file: 01_front_matter_ICLR.tex
\title{Large Language Model-Based Evolutionary \\ Optimizer: Reasoning with elitism}

\author{Shuvayan Brahmachary $^{1}$ \thanks{Equal contribution} ~ \thanks{Corresponding authors} \ ,  Subodh M. Joshi $^{1}$ $^*$, Aniruddha Panda $^{1}$, Kaushik Koneripalli $^{1}$, 
\and
{\bf Arun Kumar Sagotra $^{1}$, Harshil Patel $^{1}$, Ankush Sharma $^{1}$, Ameya D. Jagtap $^{2}$ $^\mathrm{\dagger}$, Kaushic Kalyanaraman $^{1}$} \\
\and(shuvayan.brahmachary, subodh-madhav.joshi, aniruddha.panda, 
kaushik.koneripalli, \\arun.sagotra, harshil.patel,
ankush.sharma, kaushic.kalyanaraman)@shell.com, \ ameya\textunderscore jagtap@brown.edu \\
\And
\hspace{-0.4cm} $^1$ Computational Science Group,  \ \ \ \ \ \ \ \ \ \ \ \ \ \ \ \ \ \ \ \ \ \ \ \ \ \ \ \ \ \ \  \ \ \ \ \ \ \ \ \ \ \ \ \hfill $^2$ Division of Applied Mathematics, \\
Shell India Markets Pvt. Ltd. \ \ \ \ \ \ \ \ \ \ \ \ \ \ \ \ \ \ \ \ \ \ \ \ \ \ \ \ \ \ \ \ \ \ \ \ \ \ \ \ \ \ \ \hfill  Brown University, RI 02912, USA\\
}

\iclrfinalcopy 

%% file: 02_abstract.tex
\begin{abstract}
Large Language Models (LLMs) have demonstrated remarkable reasoning abilities, prompting interest in their application as black-box optimizers. This paper asserts that LLMs possess the capability for zero-shot optimization across diverse scenarios, including multi-objective and high-dimensional problems. We introduce a novel population-based method for numerical optimization using LLMs called Language-Model-Based Evolutionary Optimizer (LEO). Our hypothesis is supported through numerical examples, spanning benchmark and industrial engineering problems such as supersonic nozzle shape optimization, heat transfer, and windfarm layout optimization. We compare our method to several gradient-based and gradient-free optimization approaches. While LLMs yield comparable results to state-of-the-art methods, their imaginative nature and propensity to hallucinate demand careful handling. We provide practical guidelines for obtaining reliable answers from LLMs and discuss method limitations and potential research directions.
\end{abstract}

%% file: 1_introduction.tex
The advent of Large Language Models (LLMs) has sparked a revolution in generative Artificial Intelligence (AI)
research \citep{RecentNLPSurvey2024, zhao2023survey, LIU2023100017}.
Since the introduction of transformer model \citep{NIPS2017_3f5ee243}, the generative AI research has seen a surge of activity and every subsequent generation of 
LLM models have become exceedingly more capable than the previous ones. 
For example, the first decoder-only model developed by OpenAI in 2018 was called
Generative Pre-Trained transformers (GPT) based on the transformer architecture, 
was capable of processing textual data alone \citep{zhao2023survey}, 
while OpenAI's latest GPT-4 model released in
2023 is multi-modal, i.e., capable of dealing with natural language, code, as well as images \citep{openai2023gpt4}. 
Several studies since then have shown that Large Language Models (LLMs), such as GPT-4, possess strong reasoning ability
\citep{huang2023reasoning, kojima2023large}.
Studies have also shown that 
a LLM's performance can be further improved by techniques such as in-context learning 
\citep{NEURIPS2020_1457c0d6}, chain-of-thought prompting \citep{wei2023chainofthought}, and tree-of-thought prompting \citep{yao2024tree, long2023large}.
 
Some examples that highlight the generalization capability of LLM models are: a) Gato \citep{reed2022generalist}, a generalist multi-modal agent based on a LLM capable of performing several tasks. b) Eureka \citep{ma2023eureka}, a human-level reward design algorithm using LLMs, is a gradient-free in-context learning approach to reinforcement learning for robotic applications. c) Voyager \citep{wang2023voyager}, a LLM-powered AI agent, has shown the ability to conduct autonomous exploration, skill acquisition, and discovery in an open-ended Minecraft world without human intervention. 
To test the reasoning and generalization ability of new generative AI models,  
Srivastava et. al. published a suite of benchmarks containing 204 problems called the Beyond the Imitation Game benchmark (BIG-bench) \citep{srivastava2023imitation}. 

This reasoning and generalization ability of LLMs has sparked interest in exploring use of LLM models as 
AI agents, particularly for applications in science and technology. 
\citet{bran2023chemcrow} developed an autonomous AI agent called ChemCrow for computational chemistry research.
This AI agent has demonstrated remarkable ability to accomplish tasks across organic 
synthesis, drug discovery, and material design. 
Similarly, \citet{boiko2023coscientist} presents an autonomous AI agent based on an LLM for chemical engineering research.
\citet{blanchard2022GAmol} use masked LLMs for automating genetic mutations for molecules for application in drug likeness and synthesizability. 
\citet{zhang2023automlgpt} present AutoML-GPT, an AI agent that acts as a bridge between various AI models as well as 
dynamically trains other AI models with optimized hyperparameters.
\citet{Stella2023} show that generative AI models can accelerate robot design process at conceptual as well as technical level. They further propose a human-AI co-design strategy for the same. 
\citet{zheng2023gpt4} explores the generative ability of GPT4 as a black-box optimizer to navigate architecture search space, making it an effective agent for neural architecture search.
\citet{singh2022progprompt} study the utility of LLMs as task planners for robotic applications.
\citet{Jablonka2024} show that a fine-tuned GPT-3 model can outperform many other ML models, particularly in the low-data limit, for predictive Chemistry.

A common thread which passes through the studies mentioned above is the 
ability of LLMs to find an optimal solution for complex multi-objective optimization problems at hand. 
This has attracted a great deal of attention from the scientific community. Several studies have been published 
exploring the ability of LLMs to work as black-box optimizers. 
\citet{melamed2023propane} presents an automatic prompt optimization framework called PROPANE. 
In a method called InstructZero, Chen et~al. optimize a low-dimensional soft-prompt to an open-source LLM, 
which in turn generates the prompt for the black-box LLM, which then performs a zero-shot evaluation  \citep{chen2023instructzero}. 
The soft prompt is optimized using a Bayesian optimization method. 
Deepmind released a general hyperparameter optimization framework called Optformer based on the transformer architecture \citep{optformer}. 
The idea of generation of optimized prompts automatically is also explored in \citep{zhou2023large}.
Similarly, \citet{pryzant2023automatic} explores incorporating
gradient descent into automatic prompt optimization. 
\citet{chen2024prompt} introduces a discrete prompt-optimization framework incorporating human-designed feedback rules.
We also see some examples of using LLM within a
Reinforcement Learning (RL) framework for optimization \citep{chen2021decision,wang2017learning,ma2023eureka}.

While the examples mentioned so far focused on optimized prompt generation, a few studies have also explored
using LLMs for mathematical optimization directly. 
\citet{liu2023large} propose LLM-based Evolutionary Algorithm (LMEA), in which 
a LLM is responsible for the selection of the parent solution, mutation, cross-over, and generation of a new solution. 
\citet{guo2023optimizing} conducts an assessment of the various optimization abilities of LLM. 
Their study concludes that LLMs can perform optimization, including gradient descent, hill-climbing, grid search 
and black-box optimization well, particularly when the sample sizes are small. 
\citet{pluhacek2023swarm} presents a strategy for using LLM for swarm intelligence-based optimization algorithms. 
\citet{liu2023largeMOO} proposes LLM-based multi-objective optimization method, where LLM serves as black-box search 
operator for Multi-Objective Evolutionary Algorithms (MOEA). 
\citet{liu2023AEL} proposes using LLMs for optimization algorithm evolution in a framework called AEL (which stands for Algorithm Evolution using LLMs).  
\citet{liu2023language} adopt 
automatic hill-climbing process using language models as black box optimizers for 
vision-language models. They show that LLMs utilize implicit gradient direction for more efficient search. 
Optimization by PROmpting (OPRO) framework from Google Deepmind generates new solutions autoregressively 
and is seen to outperform human-level prompts \citep{yang2023large}. 
We also see examples of using LLM for mathematical operations and optimization; for example, 
deep learning for symbolic mathematics \citep{lample2019deep}, LLM for symbolic regression \citep{valipour2021symbolicgpt, agarwal2021analyzing}, 
using transformers for linear algebra, including matrix operations and eigen-decomposition \citep{charton2022linear} etc.
In a framework called OptiMUS, LLM are used for formulating and solving Mixed-Integer Linear Programming (MILP)
problems \citep{ahmaditeshnizi2023optimus}.
\citet{zhang2023HO} use LLMs for hyperparameter optimization.
\citet{romera2023mathematical} introduce an evolutionary procedure called FunSearch (short for searching in function space), where a pretrained LLM model is paired with a systematic evaluator for efficient function space search.

In this paper, we focus on the ability of LLMs to perform black-box optimization. 
We adopt a population based approach and propose a novel explore-exploit policy using LLMs for generation of new samples.
We call this method a Language-model based Evolutionary Optimizer (LEO). 
The main contributions of this paper are as follows: 
\begin{enumerate}
\item We introduce a novel optimization
strategy called Language-model-based Evolutionary Optimizer (LEO). This is a population-based, parameter-free method in which a LLM is used to generate new candidate solutions and an elitist framework consisting of separate explore and exploit pools of solutions is used as guard rails. This strategy helps harness the optimization capabilities of LLMs while mitigating the risks of getting stuck in local optima. We present the details of the algorithm in Sections 1 and 2. Additionally, we highlight the rationale for adopting a population-based strategy for optimization in Section 2. 
\item We present distinguishing features of our method compared to other auto-regressive, evolutionary, or population-based methods using LLMs for black-box optimization, such as in \citep{liu2023large,yang2023large,guo2023optimizing,liu2023largeMOO} (Section 2).
\item We solve several benchmark problems in numerical optimization, single and multi-objective, as well as explore the ability of LLMs to solve high-dimensional problems. Additionally, we demonstrate the application of our method to engineering problems such as shape optimization, heat transfer, and windfarm layout optimization (Section 3).
\item We juxtapose our method against the state-of-the art methods for optimization, both gradient based and gradient-free (Section 3). 
\item We provide evidence of LLM's ability to reason and perform numerical optimization with the help of two tests (Section 4). 
\end{enumerate}
Lastly, we present our analyses and discussions, followed by a conclusion (Section 5).

%% file: 2_methodology.tex
\subsection{Rationale towards populated-based approach}
\label{llm1Dopt}
{
In this section, we provide the rationale behind the population-based approach for solving complex non-convex optimisation problems. While non-population-based or gradient-based methods are preferred for their quick turn-around time towards convergence, the final solutions are likely to get trapped in local optima for non-convex problems. In this section, we setup a quick experiment to demonstrate this idea via the LLM-assisted optimisation framework without a population-based strategy to generate a new candidate solution for every optimisation iteration. To further elaborate, the framework for this toy experiment uses LLM to generate a new candidate solution given the history of the number of previously generated solutions, $n_{\mathrm{hist}}$. This approach is strictly not a population-based strategy but rather an adaptive way to generate new candidate solutions that is likely to minimise a function value. The prompt corresponding to this test is mentioned in Table \ref{prompt_reasoning} in the Appendix. {color{blue} Throughout this work, we use the openAI GPT-3.5 Turbo 0613 model for all our evaluations. }

\begin{figure}
        \centering
        \includegraphics[width=1\textwidth]{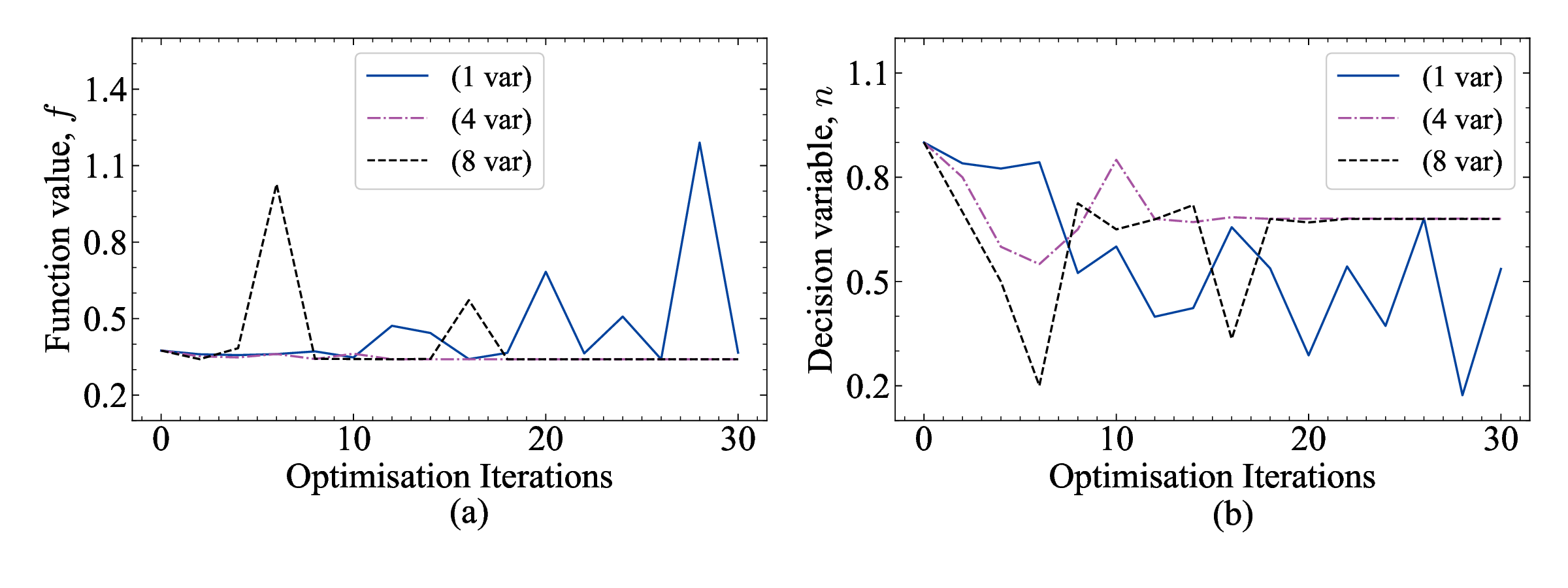}
        \caption{Convergence towards optimal nose cone shape with increasing number of decision variables as context.}
        \label{nosecone}
\end{figure}

The problem under consideration is the popular nose-cone shape optimisation. This toy problem presents itself as a single objective, single variable convex optimisation problem, where the objective is to determine the optimal decision variable (\textit{i.e.,} power-law index, $n$) that results in minimal drag coefficient body. Please refer to Section \ref{nose_cone_theory} in the appendix for further information. Figure \ref{nosecone} (a) shows the variation of the function value (\textit{i.e.,} drag coefficient) with optimisation iterations for various $n_{\mathrm{hist}}$ values, where a remarkable observation is made. It is found that the LLM struggles to find the optimal solution when $n_{\mathrm{hist}}$=1. This is not surprising and points to the lack of context in the limited number of previously generated candidate solutions. However, for higher $n_{\mathrm{hist}}$ values, the optimisation framework quickly aligns itself with the optimal value. This is also true for the corresponding decision variable value $n$ in Figure \ref{nosecone} (b). This experiment signifies that the LLM-based hybrid optimisation framework struggles to locate the optimal solution when limited context in terms of the history of previous candidate points has been  passed to it in every optimisation iteration. This is true for even a simple convex optimisation, as described above. 

\begin{figure}
        \centering
        \includegraphics[width=1\textwidth]{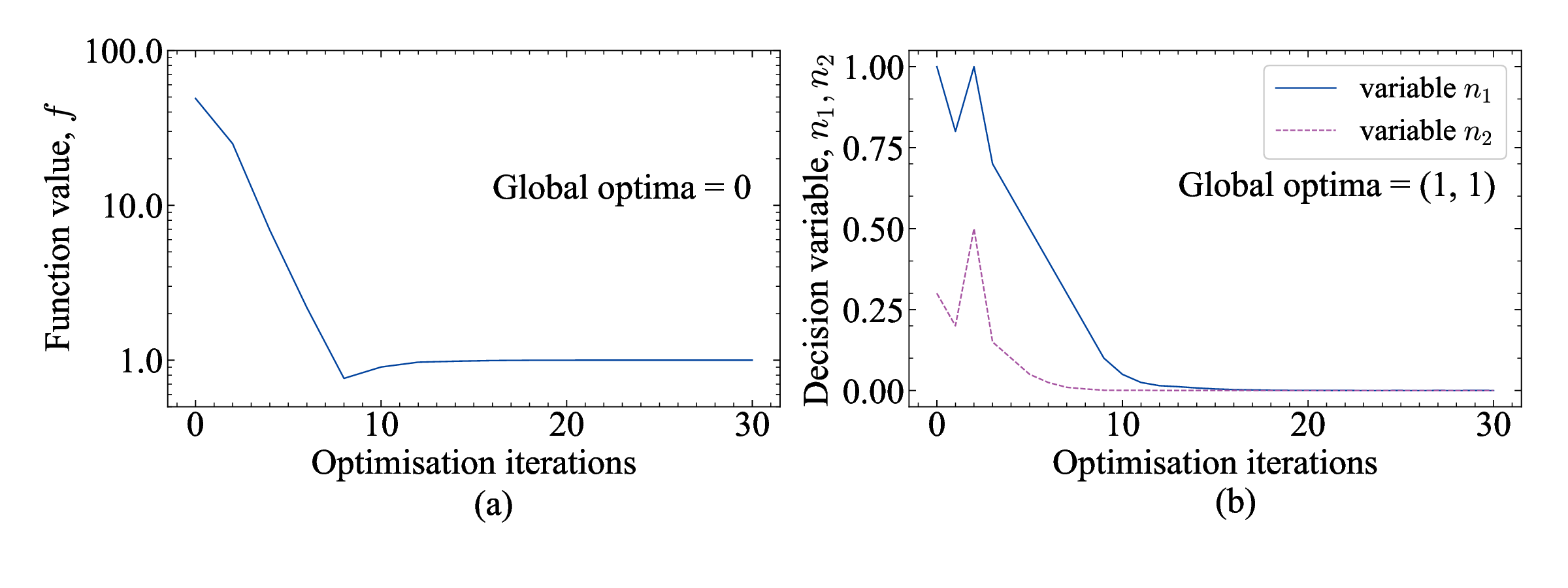}
        \caption{Performance of the LLM-based search algorithm (without a population) described in Section \ref{llm1Dopt} for the 2D Rosenbrock function.}
        \label{Rosenbrock2D}
\end{figure}

We now extend the above approach to solve a two-dimensional (2D) benchmark optimisation function, specifically the Rosenbrock function, using $n_{\mathrm{hist}}$ = 10. The global optima for this function is \textit{i.e.,} $f_{min} = 0$, corresponding to $x_{min}, y_{min} = (1,1)$. Figure \ref{Rosenbrock2D} (a) shows the variation of the objective function value, $f$, with optimisation iterations. It is clear that the hybrid approach, despite being fed the history of the last 10 candidate points, suffers from premature convergence. This can be confirmed from Fig. \ref{Rosenbrock2D}(b) where the decision variables $x, y$ are seen to converge to a local optimal solution. This failure signifies that a single trajectory of past candidate points alone is not sufficient to converge to the global optimal solution, as they might be points around the basins of a local optimal solution. 
It must be remarked that while \citet{yang2023large} allude to this phenomenon as \textit{optimisation stability} resulting from sensitivity to prompt, we are of the opinion that this is rather a misnomer and stems purely from a lack of context and richness of information from the decision variable space. Besides, our work provides a compelling demonstration of this logic.

From the above two experiments, we can see that a hybrid LLM-based optimisation approach without the rich information spanning the decision variable space (\textit{i.e.,} exploration) is unlikely to be useful for the LLM to generate good candidate points around the present candidate solutions (\textit{i.e.,} exploitation). 
While several other methods, exemplified by \citep{liu2023large, yang2023large, guo2023optimizing}, incorporate the concept of exploration versus exploitation during solution search in an auto-regressive manner, they rely on adjusting the temperature of the Language Model (LLM) to toggle between these two modes. However, this approach, although effective in simpler scenarios, has limitations. Specifically, a higher temperature introduces greater randomness into the LLM’s output, potentially hindering its ability to learn the true distribution of solutions. Conversely, an excessively low temperature can lead to mode collapse—a situation where the model consistently predicts the same (incorrect) solution. This phenomenon is particularly associated with auto-regressive LLMs, as highlighted by \citep{hopkins2023can}. Besides, the use of temperature as a user-defined parameter (unlike the self-adaptive mechanism introduced by \citep{liu2023large}) on top of an already complex optimisation method may diminish its efficiency or user friendliness. In the same vein, we will also illustrate how an existing evolutionary optimisation framework utilising user-defined parameters such as probability of crossover and probability of mutation can be simply replaced by using LLM, without the introduction of temperature. In the following section, we propose a novel approach that uses LLM to generate points that maintain a fine balance between exploration and exploitation, to arrive at the global optimal solution. This is unlike the existing approaches, where population based evolutionary optimisation strategies like genetic algorithms employ mutation and crossover to perturb the parent solution for offspring. 

\begin{figure}[!h] 
        \centering
        \includegraphics[width=0.85\textwidth]{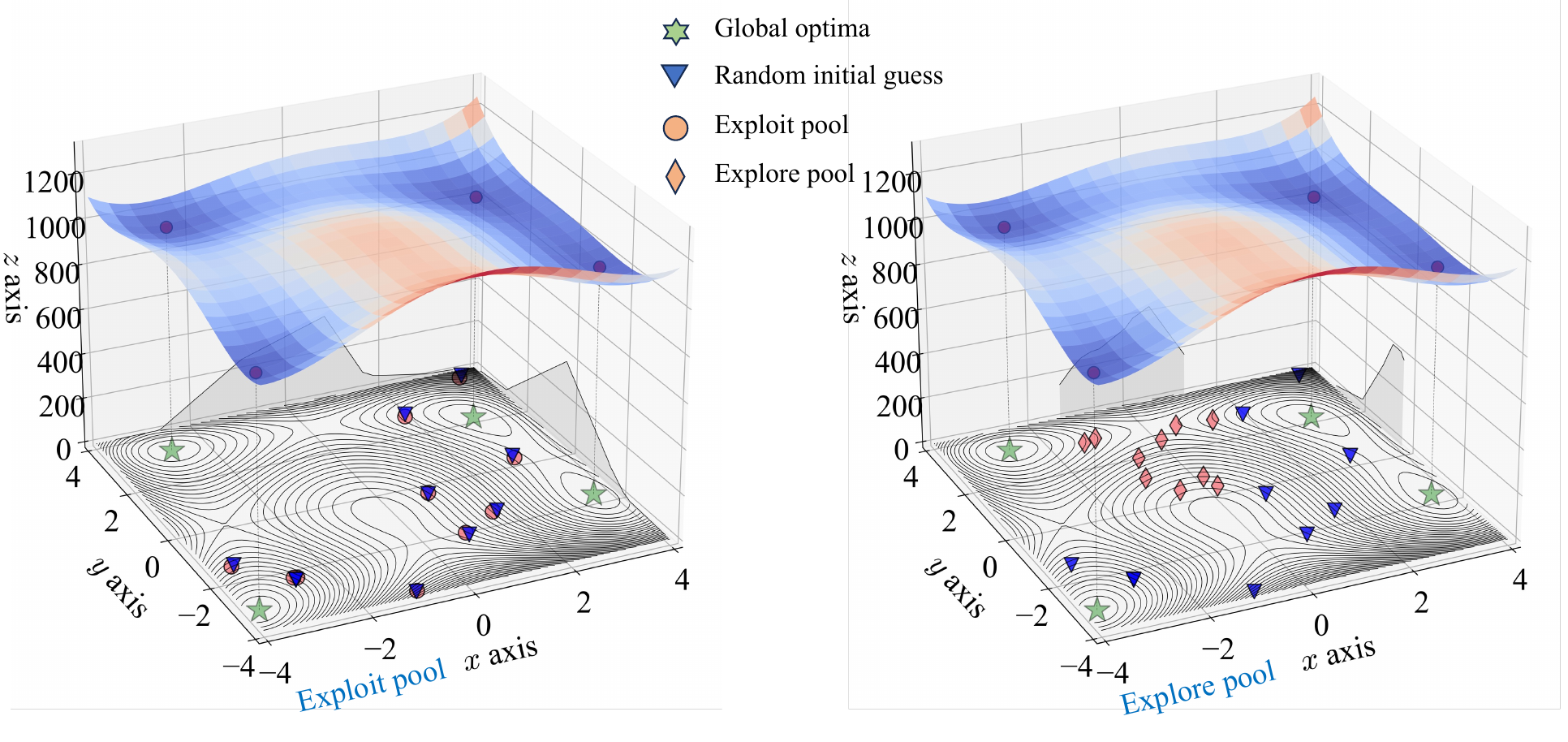}
        \caption{Schematic behind the LLM-assisted optimisation via exploration and exploitation of design space. Both figures illustrate the distribution of the initial pool of solution (triangle marker in blue color) randomly generated within the design space bounds along with the global optimal solution (in light green color). The figure(a) shows the overlay of the exploit pool of points (circle marker in sepia color) generated in close vicinity of the initial random solutions. The figure(b) shows the juxtaposition of the explore pool of solutions (diamond marker in sepia color)
        significantly away from the pool of initial randomly generated solutions. The best explore points (closest to the global optimal solution) is ported to exploit pool, following which the equal number of worst solutions from exploit pool will be removed. The $xz$ as well as $yz$ plane in both the figures depict the density distribution of the newly generated exploit and explore points.}
        \label{idea_image}
\end{figure}

\begin{figure}[!h]
        \centering
        \includegraphics[width=0.55\textwidth]{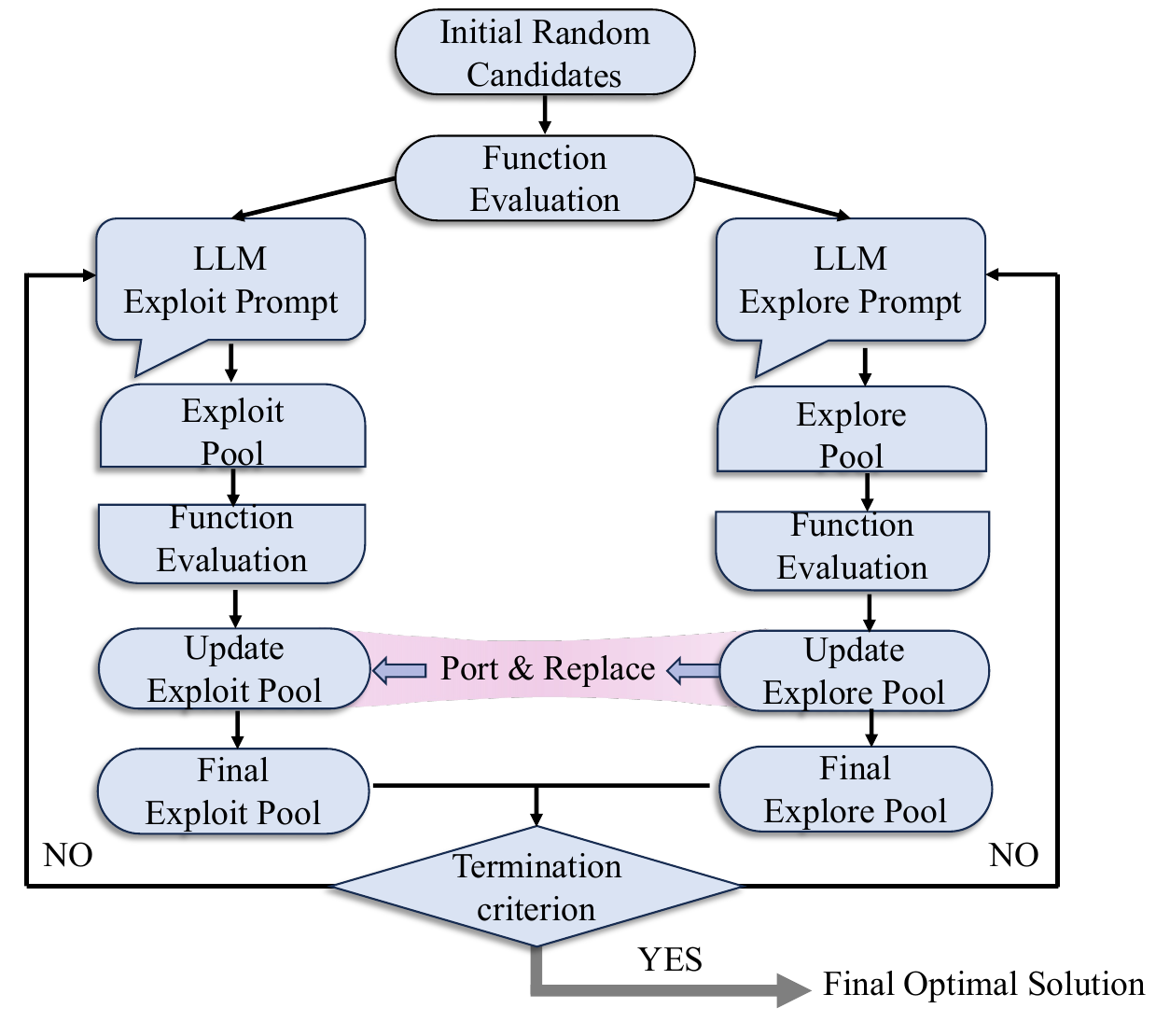}
        \caption{Schematic of the LEO framework.}
        \label{framework}
    \hfill
\label{LEO_Schematic}
\end{figure}

\subsection{\label{desc_leo} Language-model based Evolutionary Optimizer (LEO)}
In order to address the shortcomings that were discussed above, we propose a novel optimization strategy called \textbf{Language-model-based Evolutionary Optimizer (LEO)}. This approach employs LLMs along with some traditional strategies that act as `guardrails' to better guide the resulting hybrid optimization framework. Being a population-based strategy, LEO generates a pool of random initial solutions of size equal to population size $N_{pop}$, within the decision variable bounds. These initial solutions are then evaluated to determine their function values, i.e., fitness scores. At this stage, we retain all the initial pool of solutions as candidate solutions. To generate new candidate solutions that allow for a balance between exploitation and exploration, we invoke the LLM via two different prompts, \textit{i.e.,} shown in Tables \ref{exploit_prompt} and \ref{explore_prompt}, to finally create \textit{exploit} and \textit{explore} pools of candidate solutions. These pools of candidate solutions further undergo function evaluations for their respective fitness scores and are appended to the solutions from previous generations, resulting in $2N_{pop}$ solutions in each pool. At this stage, we export the \textit{$N_{port}$} best solutions from the \textit{explore} pool to the \textit{exploit} pool (and remove the \textit{$N_{port}$} worst solutions from the \textit{exploit} pool) via the \textit{Port and Filter} operation, to restrict the exploitation search to good candidate solutions with a lower function value, see Fig. \ref{idea_image}. Finally, to enforce \textit{elitism}, the $2N_{pop}$ solutions in each pool are sorted in order of their function values, and only the top $N_{pop}$ solutions are retained in each pool. This step is pictorially depicted in Fig. \ref{LEO_Schematic} and summarized in Algorithm \ref{Algo:LEOAlgorithm}. The above step is repeated until the termination criterion (\textit{i.e.,} maximum generations or optimisation iterations) is reached. 


\definecolor{}{rgb}{0.55, 0.71, 0.0}

\RestyleAlgo{ruled}
\begin{algorithm}[H]
\scriptsize
\setstretch{0.9}
\SetAlgoLined
\SetKwInput{kwInputs}{Input}
\SetKwInput{kwMinimise}{Minimise}
\SetKwInput{kwReturn}{Return}
\SetKwFunction{main}{main}
\SetKwFunction{PromptCreate}{PromptCreate}
\SetKwFunction{UpdatePopulation}{UpdatePopulation}
\SetKwFunction{PortFilter}{PortFilter}
\SetKwComment{Comment}{/* }{ */}

\SetKwProg{myalg}{Algorithm}{}{}
\myalg{\main{}}{
\nl \kwInputs{\texttt{popSize}, \texttt{maxIters}, \texttt{numVars},  \texttt{prompt\_init}, \texttt{jitter$_\mathrm{explore}$}, \texttt{jitter$_\mathrm{exploit}$}, \texttt{solToPort}, \texttt{objFun(.)}}
\nl \kwMinimise{\texttt{minFunVal}}
\nl $x_\mathrm{init}$ $\gets$ \texttt{random(popSize, numVars)}, $x_\mathrm{explore}$ $\gets$ $x_\mathrm{init}$, $x_\mathrm{exploit}$ $\gets$ $x_\mathrm{init}$ \;
\nl $y_\mathrm{init}$ $\gets$ \texttt{objFun(x$_\mathrm{init}$)}, $y_\mathrm{explore}$ $\gets$ $y_\mathrm{init}$, $y_\mathrm{exploit}$ $\gets$ $y_\mathrm{init}$ \;
\nl $iter \gets 1$ \;

\While{$iter \leq$ \texttt{maxIters}}{
    \texttt{prompt}$_\mathrm{explore}$ $\gets$ {\color{blue}\PromptCreate{\texttt{prompt}$_\mathrm{init}$, \texttt{jitter}$_{explore}$, $x_\mathrm{explore}$}} \;
    \texttt{prompt}$_\mathrm{exploit}$ $\gets$  {\color{blue}\PromptCreate{\texttt{prompt}$_\mathrm{init}$, \texttt{jitter}$_{exploit}$, $x_\mathrm{exploit}$}} \;
    $\Tilde{x}_\mathrm{explore}$, $\Tilde{x}_\mathrm{exploit}$ $\gets$ \texttt{LLM(prompt$_\mathrm{explore})$}, \texttt{LLM(prompt$_\mathrm{exploit})$} \;
    $\Tilde{y}_\mathrm{explore}$, $\Tilde{y}_\mathrm{exploit}$ = \texttt{objFun($\Tilde{x}_\mathrm{explore})$}, \texttt{objFun($\Tilde{x}_\mathrm{exploit})$} \;
    $x_\mathrm{explore}$, $y_\mathrm{explore}$ $\gets$  {\color{green!50!black}\UpdatePopulation{$x_\mathrm{explore}$, $y_\mathrm{explore}$, $\Tilde{x}_\mathrm{explore}$, $\Tilde{y}_\mathrm{explore}$}} \;
    $x_\mathrm{exploit}$, $y_\mathrm{exploit}$ $\gets$  {\color{green!50!black}\UpdatePopulation{$x_\mathrm{exploit}$, $y_\mathrm{exploit}$, $\Tilde{x}_\mathrm{exploit}$, $\Tilde{y}_\mathrm{exploit}$}} \;
     $x_\mathrm{exploit}$, $y_\mathrm{exploit}$ $\gets$  {{\color{red}\PortFilter{$x_\mathrm{explore}$, $y_\mathrm{explore}$, $x_\mathrm{exploit}$, $y_\mathrm{exploit}$, \texttt{solToPort} }}} \;
    $x_\mathrm{exploit}, y_\mathrm{exploit}$ $\gets$ \texttt{sort([$x_\mathrm{exploit}$, $y_\mathrm{exploit}$])} \;
    $x_\mathrm{explore}, y_\mathrm{explore}$ $\gets$ \texttt{sort([$x_\mathrm{explore}$, $y_\mathrm{explore}])$} \;
    $iter \gets iter + 1$
    
}
\nl \texttt{minFunVal} $\gets$ $y_\mathrm{exploit}$[0] \;
\nl \KwRet \texttt{minFunVal}\
}{}
\setcounter{AlgoLine}{0}

\SetKwProg{myproc}{Procedure}{}{}
\myproc{\color{blue}{\PromptCreate{\texttt{prompt}, \texttt{jitter}, $x$}}}{
\nl $x_\mathrm{jitter} \gets x +$ \texttt{jitter}  \;
\nl \texttt{prompt}$_\mathrm{update} \gets \text{Insert } x_\mathrm{jitter} \text{ in }$ \texttt{prompt} \;
\nl \KwRet \texttt{prompt}$_\mathrm{update}$\;
}
\setcounter{AlgoLine}{0}

\SetKwProg{myproc}{Procedure}{}{}
\myproc{{\color{green!50!black}\UpdatePopulation{$x$, $y$, $\Tilde{x}$, $\Tilde{y}$}}}{
\nl $x \gets$ \texttt{concat}($x, \Tilde{x}$) \;
\nl $y \gets$ \texttt{concat}($y, \Tilde{y}$) \;
\nl \KwRet $x$, $y$ \;
}
\setcounter{AlgoLine}{0}

\SetKwProg{myproc}{Procedure}{}{}
\myproc{{\color{red}\PortFilter{$x_\mathrm{explore}$, $y_\mathrm{explore}$, $x_\mathrm{exploit}$, $y_\mathrm{exploit}$, \texttt{numPort}}}}{
\nl $x_\mathrm{top}$, $y_\mathrm{top}$ $\gets$ \texttt{sort([$x_\mathrm{explore}$, $y_\mathrm{explore}$])}[0:\texttt{numPort}] \Comment*[r]{Sort w.r.t $y_\mathrm{explore}$}
\nl $x_\mathrm{exploit}$, $y_\mathrm{exploit}$ $\gets$ \texttt{sort([$x_\mathrm{exploit}$, $y_\mathrm{exploit}$])} \Comment*[r]{Sort w.r.t $y_\mathrm{exploit}$}
\nl $x_\mathrm{exploit}[-\texttt{numPort}:]$ $\gets$ $x_\mathrm{top}$ \;
\nl $y_\mathrm{exploit}[-\texttt{numPort}:]$ $\gets$ $y_\mathrm{top}$ \;
\nl \KwRet $x_\mathrm{exploit}$, $y_\mathrm{exploit}$\; 
}
\caption{LEO algorithm.}
\label{Algo:LEOAlgorithm}
\end{algorithm}

%% file: 3_numerical_results.tex

In this section, we evaluate our proposed optimisation strategy via a series of test cases, ranging from simple benchmark problems to engineering applications. These tests are classified into four categories: (a) simple benchmark function optimisation problems; (b) multi-objective optimization problems; (c) high-dimensional benchmark optimization problems; and (d) industry-relevant engineering optimization problems. This allows us to scrutinise the approach for a range of problems of different complexity. 

\subsection{Benchmark optimisation functions}

We begin our experiments by recognising the need to evaluate the merits of LEO on a range of test cases against established optimisation methods. Consequently, in this section, we compare the proposed optimization algorithm (LEO) with various gradient-based and gradient-free optimisation algorithms for 6 different 2D benchmark optimisation problems. The corresponding equations describing the function, decision variable bounds, as well as the location of optima are shown in Table \ref{Table:2DBenchmarkFunctions}, for all the respective 2D benchmark functions. 

\begin{table}[]
\centering
\resizebox{\textwidth}{!}{%
\begin{tabular}{|l|l|l|l|}
\hline
\begin{tabular}[c]{@{}l@{}}2D Benchmark \\ Objective \\ Functions \end{tabular} & Equation & Global Minima & Domain \\ \hline
ScaledSphere & $f(x,y) = x^{2}+y^{4}$ & $f(0.,0.) = 0.$ & $-1 \leq x,y \leq 4$ \\ \hline
Himmelblau & $f(x,y) = (x^{2}+y-11)^2 + (x+y^2-7)^2$ & \begin{tabular}[c]{@{}l@{}} $f(3.0,2.0)=0.0$ \\ $f(-2.805118,3.131312)=0$ \\ $f(-3.779310,-3.283186)=0$ \\ $f(3.584428,-1.848126)=0$ \end{tabular} & $-5 \leq x,y \leq 5$ \\ \hline
Rosenbrock2D & $f(x,y) = 100(y-x^{2})^{2} + (1-x)^{2}$ & $f(1.,1.)=0.$ & $-\infty \leq x,y \leq \infty $ \\ \hline
Sphere2D &  $f(x,y)$ = $x^{2} + y^{2}$ & $f(0.,0.)=0.$ & $-\infty \leq x,y \leq \infty $  \\ \hline
Beale & \begin{tabular}[c]{@{}l@{}} $f(x,y) = (1.5-x+xy)^{2}$ + \\ $(2.25-x+xy^{2})^{2} + (2.625-x+xy^{3})^{2}$\end{tabular} & $f(3.,0.5)=0.$ & $-4.5 \leq x,y \leq 4.5 $ \\ \hline
Goldstein Price & \begin{tabular}[c]{@{}l@{}}$f(x,y) = \left[ 1+(x+y+1)^{2}(19-14x+3x^{2}-14y+6xy+3y^{2})\right]$ \\ $\left[ 30 + (2x-3y)^{2}(18-32x+12x^{2}+48y-36xy+27y^{2})\right]$\end{tabular} & $f(0.,-1)=3.$ & $-2 \leq x,y \leq 2 $ \\ \hline
\end{tabular}
}
\caption{\label{Table:BenchmarkFunctions} 2D Benchmark test functions with their equations, domain, and global minima.}
\label{Table:2DBenchmarkFunctions}
\end{table}

Among the gradient-based optimization algorithms, we choose Stochastic Gradient Descent (SGD) \cite{sgd}, Adam \cite{adam}, and L-BFGS-B \cite{lbfgs} algorithms, whereas among the gradient-free optimization algorithms, we employ Simulated Annealing (SA) \cite{simulatedAnnealing}, Covariance matrix adaptation evolution strategy, CMA-ES \cite{cma-es}, and Constrained Optimization by Linear Approximation, COBYLA \cite{cobyla} to compare against LEO. All the corresponding results are shown in Table \ref{Table:2DBenchmarkTraditionalAlgos}. The reported values are the median values from 100 independently seeded runs, each corresponding to the obtained minima at the end of 1000 function evaluations.  It is found that L-BFGS-B and CMA-ES are the best performing SOTA gradient-based and gradient-free optimisation methods, respectively achieving the global optima in all the benchmark optimisation functions. However, it is interesting to note that LEO also attains the global minima, albeit with a slightly lower precision when compared across the 100 independent runs. Nevertheless, LEO seems to perform almost at par with the established SOTA methods like L-BFGS and CMA-ES while outperforming methods like SGD, SA, COBYLA and even Adam, especially on the Goldstein-Price problem, which happens to be the trickiest one to solve.

\begin{table}[]
\centering
\resizebox{\textwidth}{!}{%
\begin{tabular}{|c|c|ccccccc|ccccc|c|}
\hline
 &  & \multicolumn{7}{c|}{Traditional Gradient   based methods} & \multicolumn{5}{c|}{Traditional Gradient   Free methods} &  \\ \cline{3-14}
 &  & \multicolumn{1}{c|}{} & \multicolumn{2}{c|}{\begin{tabular}[c]{@{}c@{}}SGD with \\ momentum\end{tabular}} & \multicolumn{3}{c|}{\begin{tabular}[c]{@{}c@{}}Adam \\ ($\beta_{2}$ = 0.999)\end{tabular}} &  & \multicolumn{3}{c|}{\begin{tabular}[c]{@{}c@{}}Simulated Annealing with \\ exponential temperature decay\end{tabular}} & \multicolumn{1}{c|}{} &  &  \\ \cline{4-8} \cline{10-12}
\multirow{-3}{*}{\begin{tabular}[c]{@{}c@{}}2D Benchmark \\ Objective \\ Functions\end{tabular}} & \multirow{-3}{*}{\begin{tabular}[c]{@{}c@{}}Global \\ minima\end{tabular}} & \multicolumn{1}{c|}{\multirow{-2}{*}{SGD}} & \multicolumn{1}{c|}{$\nu$=0.5} & \multicolumn{1}{c|}{$\nu$=0.9} & \multicolumn{1}{c|}{{\color[HTML]{0D0D0D} $\beta_{1}$ = 0.1}} & \multicolumn{1}{c|}{{\color[HTML]{0D0D0D} $\beta_{1}$ = 0.5}} & \multicolumn{1}{c|}{{\color[HTML]{0D0D0D} $\beta_{1}$ =  0.9}} & \multirow{-2}{*}{L-BFGS-B} & \multicolumn{1}{c|}{T = 0.1} & \multicolumn{1}{c|}{T = 1} & \multicolumn{1}{c|}{T = 10} & \multicolumn{1}{c|}{\multirow{-2}{*}{CMA-ES}} & \multirow{-2}{*}{COBYLA} & \multirow{-3}{*}{LEO (ours)} \\ \hline
ScaledSphere & 0.0000 & \multicolumn{1}{c|}{0.0039} & \multicolumn{1}{c|}{0.0008} & \multicolumn{1}{c|}{0.0000} & \multicolumn{1}{c|}{0.0000} & \multicolumn{1}{c|}{0.0000} & \multicolumn{1}{c|}{0.0000} & 0.0000 & \multicolumn{1}{c|}{0.0000} & \multicolumn{1}{c|}{0.0000} & \multicolumn{1}{c|}{0.0000} & \multicolumn{1}{c|}{0.0000} & 0.0000 & 0.0000 \\ \hline
Himmelblau & 0.0000 & \multicolumn{1}{c|}{0.0000} & \multicolumn{1}{c|}{0.0000} & \multicolumn{1}{c|}{0.0000} & \multicolumn{1}{c|}{0.0000} & \multicolumn{1}{c|}{0.0000} & \multicolumn{1}{c|}{0.0000} & 0.0000 & \multicolumn{1}{c|}{0.0003} & \multicolumn{1}{c|}{0.0003} & \multicolumn{1}{c|}{0.0002} & \multicolumn{1}{c|}{0.0000} & 0.0000 & 0.0086 \\ \hline
Rosenbrock & 0.0000 & \multicolumn{1}{c|}{0.0362} & \multicolumn{1}{c|}{0.0479} & \multicolumn{1}{c|}{0.0406} & \multicolumn{1}{c|}{0.0986} & \multicolumn{1}{c|}{0.0192} & \multicolumn{1}{c|}{0.0155} & 0.0000 & \multicolumn{1}{c|}{0.0001} & \multicolumn{1}{c|}{0.0002} & \multicolumn{1}{c|}{0.0005} & \multicolumn{1}{c|}{0.0000} & 0.0965 & 0.0012 \\ \hline
Sphere & 0.0000 & \multicolumn{1}{c|}{0.0000} & \multicolumn{1}{c|}{0.0000} & \multicolumn{1}{c|}{0.0000} & \multicolumn{1}{c|}{0.0000} & \multicolumn{1}{c|}{0.0000} & \multicolumn{1}{c|}{0.0000} & 0.0000 & \multicolumn{1}{c|}{0.0000} & \multicolumn{1}{c|}{0.0000} & \multicolumn{1}{c|}{0.0000} & \multicolumn{1}{c|}{0.0000} & 0.0000 & 0.0000 \\ \hline
Beale & 0.0000 & \multicolumn{1}{c|}{0.1330} & \multicolumn{1}{c|}{3.2964} & \multicolumn{1}{c|}{4.6201} & \multicolumn{1}{c|}{1.1810} & \multicolumn{1}{c|}{0.1995} & \multicolumn{1}{c|}{0.9322} & 0.0000 & \multicolumn{1}{c|}{0.0000} & \multicolumn{1}{c|}{0.0000} & \multicolumn{1}{c|}{0.0002} & \multicolumn{1}{c|}{0.0000} & 0.0071 & 0.0156 \\ \hline
GoldsteinPrice & 3.0000 & \multicolumn{1}{c|}{72.5447} & \multicolumn{1}{c|}{54.4256} & \multicolumn{1}{c|}{30.3447} & \multicolumn{1}{c|}{39.4401} & \multicolumn{1}{c|}{30.0070} & \multicolumn{1}{c|}{30.0000} & 3.0000 & \multicolumn{1}{c|}{30.0038} & \multicolumn{1}{c|}{30.0047} & \multicolumn{1}{c|}{30.0044} & \multicolumn{1}{c|}{3.0000} & 60.1236 & 3.0023 \\ \hline
\end{tabular}%
}
\caption{Comparison of proposed algorithm with traditionally used Gradient-based as well as Gradient-free methods. All evaluations were continued until the termination criterion of 1000 function evaluations}
\label{Table:2DBenchmarkTraditionalAlgos}
\end{table}

\begin{figure}[!htbp] 
    \centering
    \includegraphics[width=\textwidth, height=\textheight,keepaspectratio]{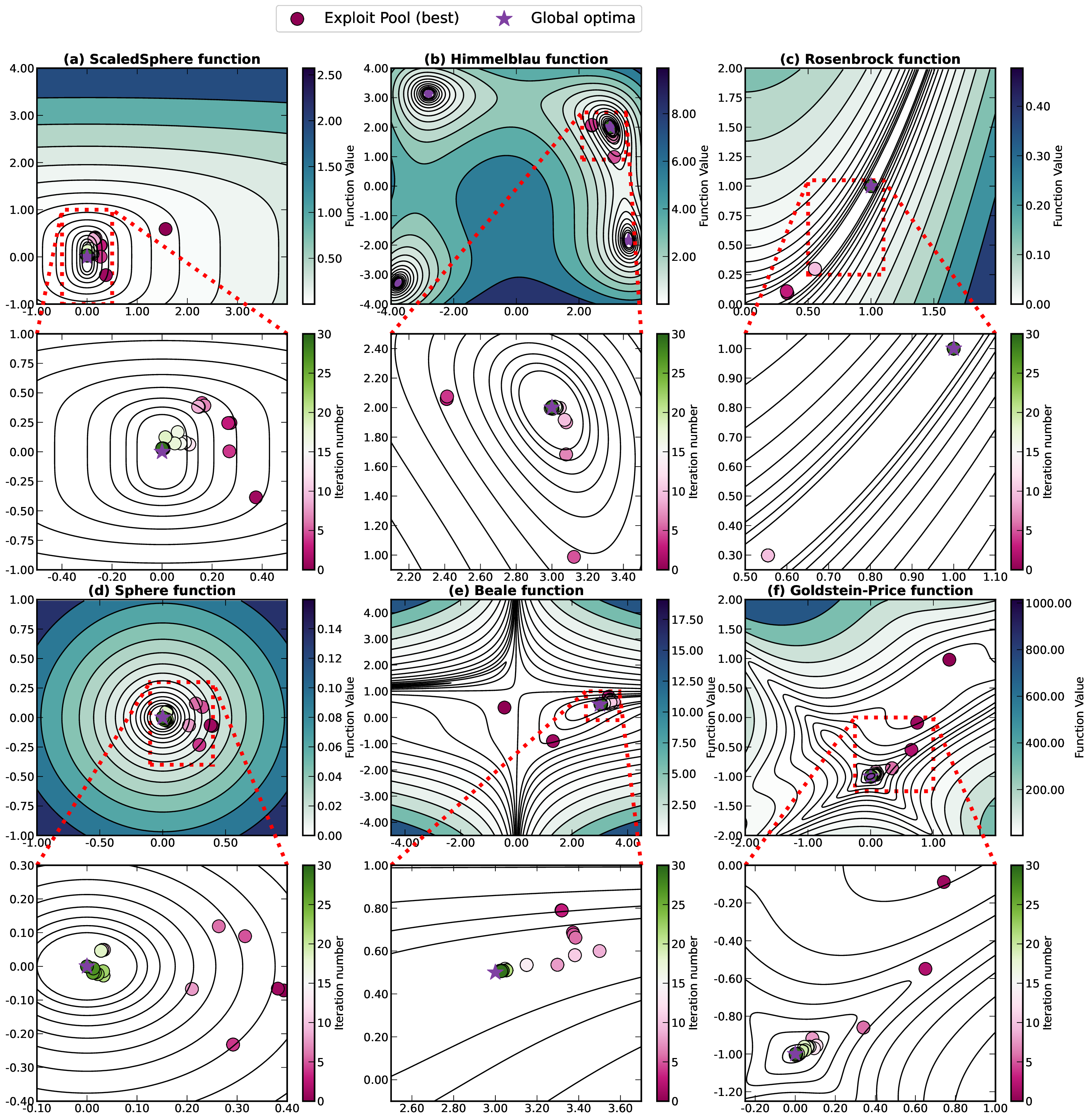}
    \caption{2D benchmark functions and their convergence plots using LEO optimization.}
    \label{Fig:2DBenchmarkConvergencePlots}
\end{figure}

A visual representation of the convergence using one representative experiments among available 100 is depicted by the solution with the lowest function value in exploit pool $f_\mathrm{min}$, obtained using LEO on the six 2D benchmark optimisation functions is shown in Figure \ref{Fig:2DBenchmarkConvergencePlots}. Every solution is color coded based on the optimisation iteration at which it was generated. It is found that while sampling of random initial solution results in $f_\mathrm{min}$ being quite distant from the respective global optimal solutions, as optimisation progresses, LEO is able to quickly navigate towards the global optimal solution. It must be noted that the convergence rate is rather high at the start of the optimisation and later tapers off near the global optimal solution depending upon the topology of the functional value space (will be quantitatively demonstrated later in following sections). This is indicative of the fact that the rate of convergence at the start is heavily influenced by solutions obtained from exploration pool while towards the end, the success of the approach is largely determined by the solutions obtained from exploitation pool. With these promising observations, we now shift our attention towards much more complex problems in the following sections. 


\textbf{Computational cost:} 

To enable a fair comparison across multiple gradient-based and gradient-free optimisation methods, it is imperative to consider the number of function evaluations while keeping the total number of iterations constant. This necessitates treating the total number of optimisation iterations as a variable. Consider the case where the gradient descent method for a problem with $n_\mathrm{var}$ variables requires $n_\mathrm{var}$+1 number of function evaluations per optimisation iteration compared to a population-based approach with population size $n_\mathrm{pop}$ that warrants $n_\mathrm{pop}$ number of function evaluations per optimisation iterations. Considering this, in the present work, we choose to keep the same number of function evaluation calls (i.e., 1000), or 30 optimisation iterations for LEO.

For the above set of experiments reported in Table \ref{Table:2DBenchmarkTraditionalAlgos}, optimization algorithms like CMA-ES and LBFGS are extremely fast. For instance, 100 randomly seeded experiments take less than a minute across all 6 benchmark functions. On the contrary, LEO, however, takes approximately 3 hours for all six functions to be repeated using 100 random seeds. We attribute this high computational time by LEO to predominantly 2 factors. Firstly, time consumed by the LLM function call: from our experiments, each call to the LLM with a $\sim$ 1000 token prompt takes anywhere between 15-20 seconds for a response from the OpenAI servers. This is already quite time-prohibitive. Secondly, due to lack of open-sourced and robust LLM models, unlike GPT-3.5 Turbo: there is a lot of growing literature on improving inference and training times of LLMs with ideas like FlashAttention \citep{dao2022flashattention, dao2023flashattention2}, model quantization \citep{lin2023awq, xiao2023smoothquant}, etc. In order to apply these ideas, one needs to have access to the model weights. Currently, our present work is limited to the use of the GPT-3.5 Turbo-0613 model because of its superior and robust performance compared to other open-sourced models available.

\subsection{Multi-objective optimization problem}

In the previous section, we demonstrated the merits LEO in tackling single objective optimisation problems for 2D standard benchmark optimisation functions. In this section, we further explore the capabilities of LEO in tackling multi-objective optimisation problems. The objective of this section is twofold: (a) Demonstrate that LLM can be used as \textit{plug-and-play} or \textit{modular} feature in the existing state-of-the art evolutionary optimisation approaches, \textit{e.g.,} non-dominated sorting algorithm (NSGA) II \citep{deb2002fast}, (b) demonstrate the ability of the resulting framework (LEO-modular) to minimize multiple conflicting objectives, which results in a distinct Pareto optimal front. 

\begin{figure}[ht]
    \centering
    \begin{subfigure}[b]{0.35\textwidth}
        \centering
        \includegraphics[width=\textwidth]{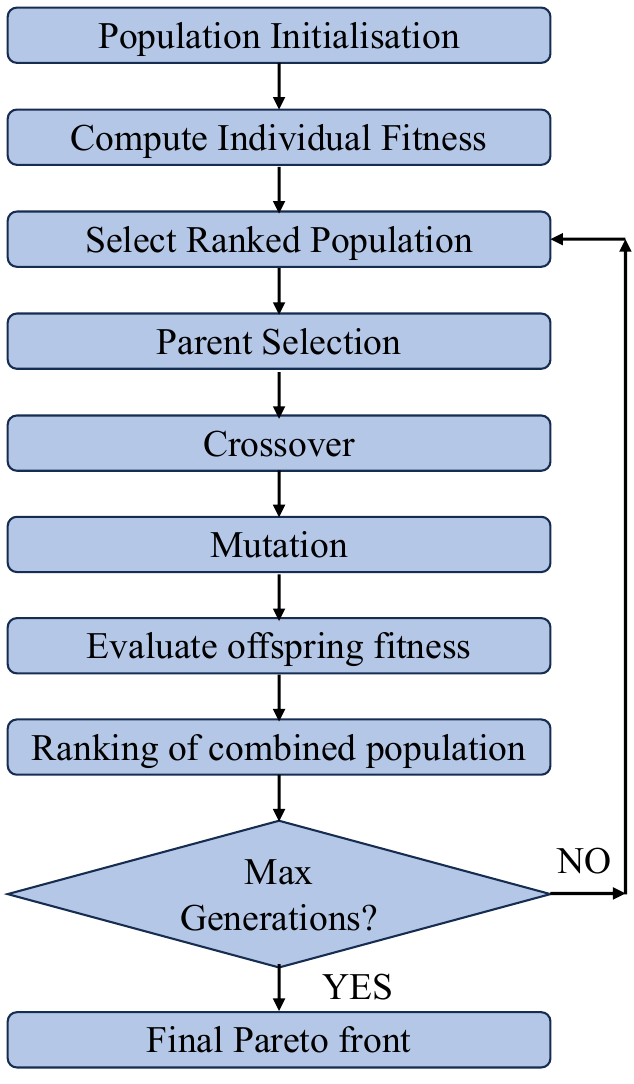}
        \caption{Schematic of the NSGA II \cite{deb2002fast} framework}
        \label{NSGA}
    \end{subfigure}%
\ \ \ \ \ \ \ \ \ 
    \begin{subfigure}[b]{0.35\textwidth}  
        \centering 
        \includegraphics[width=\textwidth]{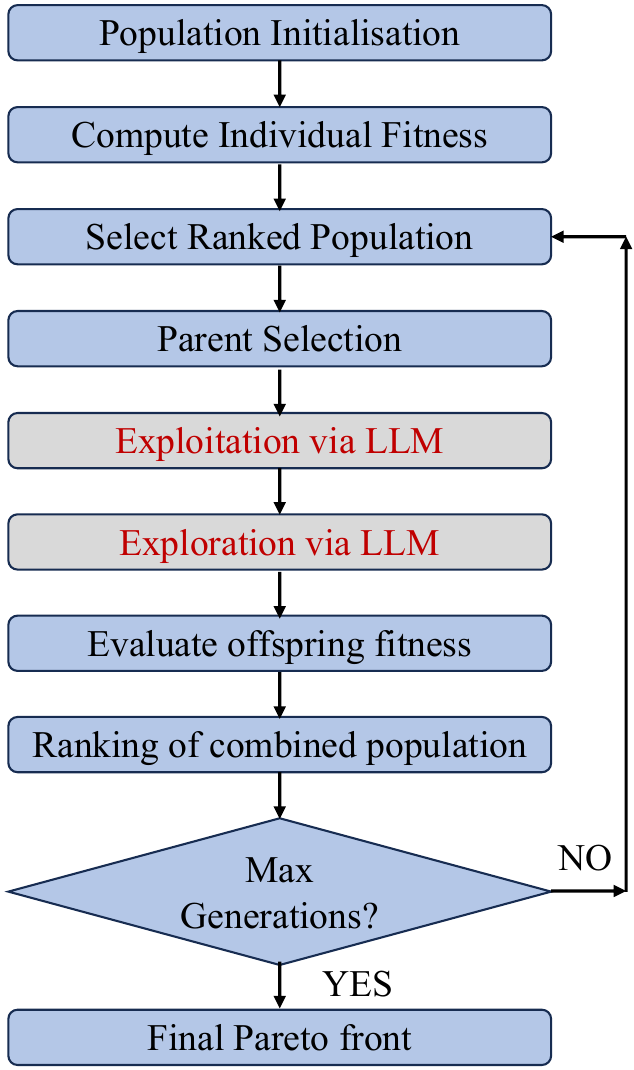}
        \caption{Schematic of the LLM assisted NSGA II framework}
        \label{NSGA_LLM}
    \end{subfigure}
    \caption{Original and modified NSGAII framework using LEO-modular}
\end{figure}

The difference between the original NSGA II and the proposed \textit{modular} feature of the LLM assisted NSGA II algorithm (or LEO-modular) can be seen from the juxtaposition of flowcharts in Figure \ref{NSGA} and Figure \ref{NSGA_LLM}. The key difference lies in the use of LLM to generate candidate solutions that explore and exploit the design space via prompts, as shown in Table \ref{prompt_NSGA}, as opposed to the traditional use of crossover and mutations. The crossover parameter employed in NSGA II algorithm generates offspring by switching parts of the same chromosome among two parents. This operation retains the best of both parents chromosomes and often leads to candidate solutions or offsprings in the near vicinity to the parents solutions, encouraging exploitation. The mutation operator in binary coded genetic algorithm randomly flips the binary encoded bits of chromosomes based on certain mutation probability. This often results in significant jump in the parent solution, thus promoting exploration. The numerical experiments using LEO in the previous section has already demonstrated evidence of generating solutions that exhibits attributes of exploration and exploitation. We now demonstrate proof of concept that LLM based exploration and exploitation works equally well within the NSGA II paradigm.

\begin{table}
\centering
\caption{Equations and Complexities of ZDT1 and ZDT3 Functions}
\label{tab:zdt_functions}
\begin{adjustbox}{width=\columnwidth,center}
\begin{tabular}{ll}
\hline
\textbf{Function} & \textbf{Complexity} \\
\hline
\textbf{ZDT1:} & Non-uniformity in the search space and a convex Pareto-optimal front. \\
$f_1(x) = x_1$ & \\
$f_2(x) = g(1 - \sqrt{f_1/g})$ & Requires algorithms to maintain diversity. \\
$g = 1 + \frac{9}{n-1}\sum_{i=2}^{n}x_i$ & \\
\hline
\textbf{ZDT3:} & Disconnected Pareto-optimal front segments introduce additional complexity. \\
$f_1(x) = x_1$ & \\
$f_2(x) = g(1 - \sqrt{f_1/g} - (f_1/g)\sin(10\pi f_1))$ & Maintaining population diversity across gaps is challenging. \\
$g = 1 + \frac{9}{n-1}\sum_{i=2}^{n}x_i$ & \\
\hline
\end{tabular}
\label{ZDT_tests}
\end{adjustbox}
\end{table}

In order to fit the scope of the original NSGA II algorithm for exploitation/exploration, we generate two offsprings using LLM via two parents as context at a given time. This must be contrasted with the approach described in Section \ref{desc_leo} (Figure \ref{LEO_Schematic}), where we generate the entire population of solutions at a time. To demonstrate the merits of the present approach we undertake two test cases named after the researchers of the seminal work \citep{ZDT_test}: ZDT1 and ZDT3. The equations describing the multiple objective functions $f_1, f_2$ are listed in Table \ref{ZDT_tests}, while also highlighting the complexity associated with the two functions. We employ these test experiments as a two-variable problem, \textit{i.e.,} $n$=2, in Table \ref{ZDT_tests}. The ZDT1 test was undertaken using a population size of 10, whereas the ZDT3 test was performed using a population size of 30. Higher population size for the ZDT3 test allows for the crisp capture of the segmented Pareto optimal front. 


\begin{figure}[ht]
    \centering 
        \includegraphics[width=0.9\textwidth]{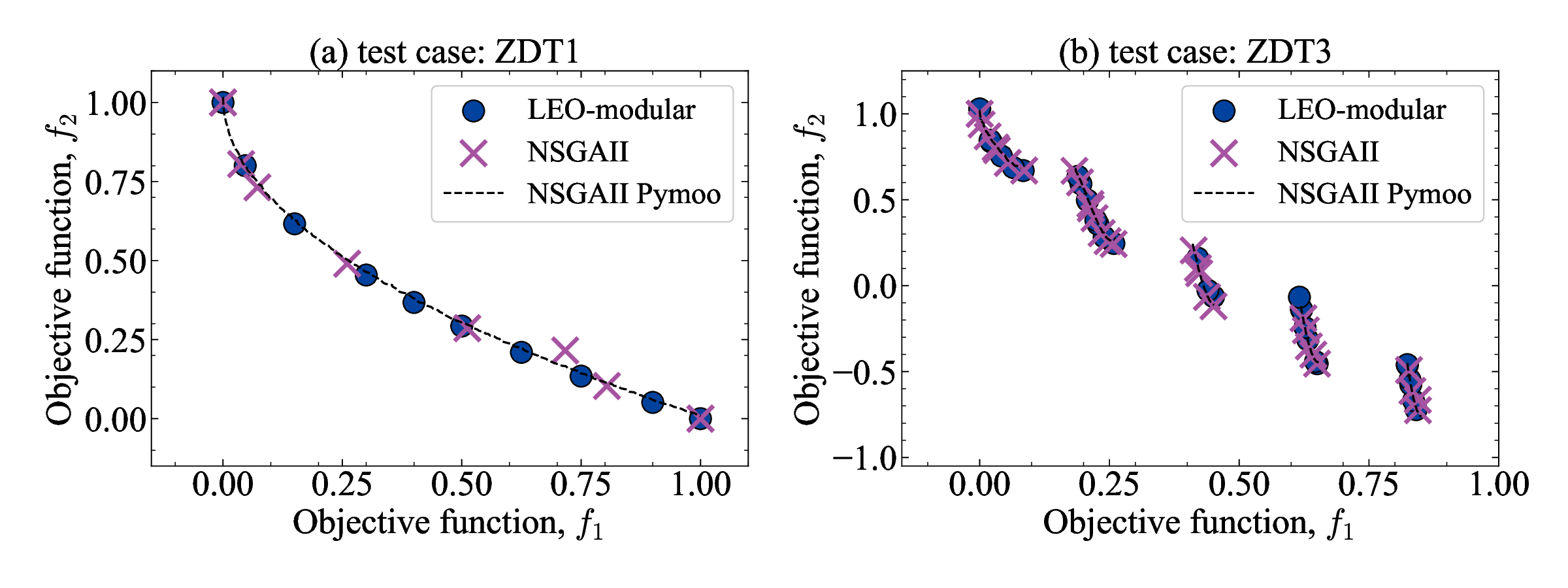}
    \caption{Comparison of the Pareto optimal front obtained from the present approach with the NSGA II algorithm }
    \label{ZDT1_ZDT3_test}
\end{figure}

Figure \ref{ZDT1_ZDT3_test}(a) shows the final Pareto optimal front obtained from the LLM-modular, along with suitable comparisons made using the original NSGA II algorithm. Every input parameter including population size, objective functions, number of optimisation iterations etc., are kept same for a fair comparison. Both these test cases were run upto a total of 40 generations or optimisation iterations. For the sake of comparison, we also present the solutions obtained from yet another implementation of the NSGA II algorithm using the open-source Pymoo \cite{pymoo} library. Pymoo-based solutions were generated using a much larger population size $n_\mathrm{pop}$ = 200, and the optimisation was carried out for a total of 100 generations. Consequently, the solutions would exhibit a dense Pareto optimal front. It can be seen that the Pareto optimal front obtained from LEO-modular is in agreement with the Pareto optimal front obtained from the traditional NSGA II algorithm. Figure \ref{ZDT1_ZDT3_test}(b) shows the Pareto optimal front for the ZDT3 test case, where a discontinuous front has been obtained. It is again found that the solutions from both approaches are in excellent agreement despite the presence of discontinuities. These observations have a remarkable significance that the LLM-based exploration and exploitation without any temperature-based parameter is able to generate solutions that distinctly identify the Pareto optimal front using. This is especially significant considering that a low population size and relatively low number of generations were employed to get the final Pareto optimal front. In the view of the authors, this is the first ever instance where LLM have been used to perturb parent solutions to generate offsprings purely based on prompts and should be viewed differently than using LLM to generate offsprings via crossover and mutation. 

\subsection{High-dimensional problem }
\label{high_dim_results}

We now extend the capabilities of LEO for high-dimensional problems. For this, we consider the Rosenbrock function which has a generalized N-dimensional formulation (2D formulation shown in Table \ref{Table:BenchmarkFunctions}). We linearly vary the dimensionality of the problem and make some interesting observations about the performance of LEO. The Rosenbrock N-dimensional function is given by:
\begin{equation}
f(\textbf{x}) = f(x_0, x_1,...,x_{n-1})  = \sum_{i=0}^{n-1} \left[ 100(x_{i+1}-x_{i}^{2})^{2} + (1-x_{i})^{2} \right]  , \hspace{4mm} -\infty < x_{i} < \infty, \forall i
\end{equation}
This function has a global minimum of $0$ as follows: $$f(\underbrace{1,...,1}_\text{n times}) = 0$$

We perform experiments by varying $n$ from 2 all the way up to 25 dimensions, giving total $7$ instances of the functions with increasing dimensionality, \textit{i.e.} $2,4,6,8,10,20,25$. Each optimization problem is run for 100 optimization iterations, and we conduct 30 such independent runs for each problem to get statistically significant results. 

A key aspect to this problem is the fact that the global optima for the Rosenbrock function, $f_{min}=0$, is at the location $(1,...,1)$. While LEO is able to accurately predict this optima, it can be argued that the number $1$ could be one of the most common numbers encountered by the LLM during it's training phase, causing the LLM to predict this number simply because of its training history. This may result in the LLM predicting the correct solution by the virtue of it being a `trivial' solution. In order to address this issue, we modify the Rosenbrock function, to have the minima shifted to an arbitrary location. The new shifted function is given by:

\begin{equation}
f(\textbf{x}) = f(x_0, x_1,...,x_{n-1})  = \sum_{i=0}^{n-1} \left[ 100((x_{i+1}-a)-(x_{i}-a)^{2})^{2} + (1-(x_{i}-a))^{2} \right]  , -\infty < x_{i} < \infty, \forall i
\end{equation}
This function has a global minimum of $0$ as follows: $$f(\underbrace{1+a,...,1+a}_\text{n times}) = 0$$

with $a$ randomly selected as $0.2913$. It is critical to point out that doing the above exercise, naturally forces LEO to thoroughly search for for better candidate solutions instead of returning a trivial solution.

\begin{figure}[ht]
    \centering 
    \includegraphics[width=0.8\textwidth]{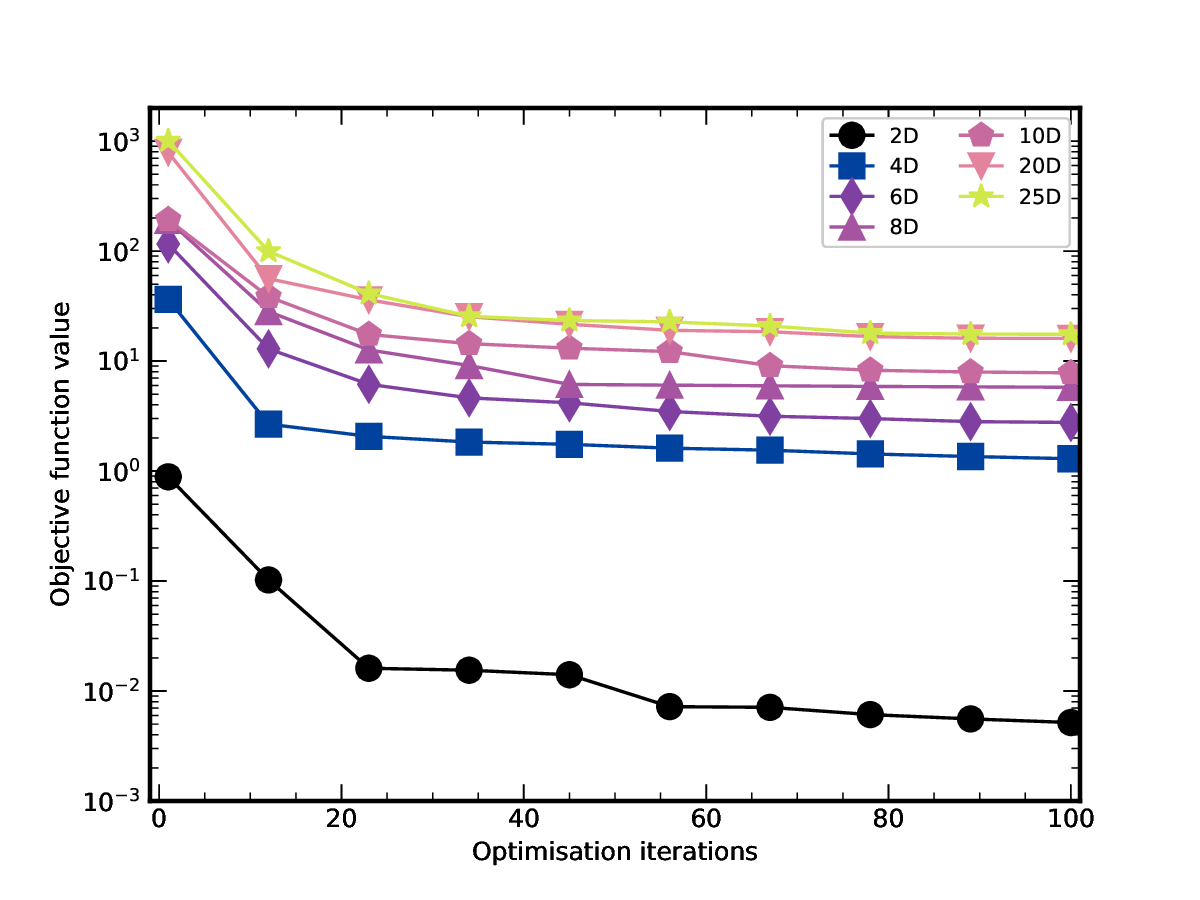}
    \caption{Convergence pattern in the $f_{min}$ obtained from exploit pool for the Rosenbrock function with different dimensions}
    \label{Fig:RosenbrockHD}
\end{figure}

Figure \ref{Fig:RosenbrockHD} shows a distribution of the mean objective function value in the exploit pool plotted with respect to the number of optimization iterations for all the $7$ instances of the functions. We notice that as the dimensionality increases, the obtained minima of the function value also increases. This is not surprising and rather aligns well with the intuition that with higher dimensionality of the problem, the population size as well as the number of optimisation iterations should also be increased. Nevertheless, the intention behind performing this exercise is to convey the consistency of the convergence rate for high-dimensional problem when limited to certain number of optimisation iterations.

\subsection{Engineering applications}
In this section, we examine the application of our method using practical examples of industrial relevance. We have selected three applications for this purpose: (a) Nozzle shape optimization for supersonic flow (b) $1D$ steady-state heat transfer problem; (c) wind-farm layout optimization. In the following subsections, we discuss the results obtained with LEO. The details of each of the problem setups are provided in the Appendix.

\subsubsection{Nozzle shape optimization}

We begin by showcasing the utility of LEO for industrial-scale optimisation problems, specifically the nozzle shape design optimisation problem. In the context of the present work, we aim to obtain a nozzle contour represented by a cubic Bezier curve that allows for maximum thrust by reducing the radial velocity component at the exit. The length, inlet radii, and outlet radii are kept constant, resulting in an effective 4-dimensional (4D) problem. For details, please refer to \ref{nozzle_theory}.

\begin{figure}
        \centering
        \includegraphics[width=1\textwidth]{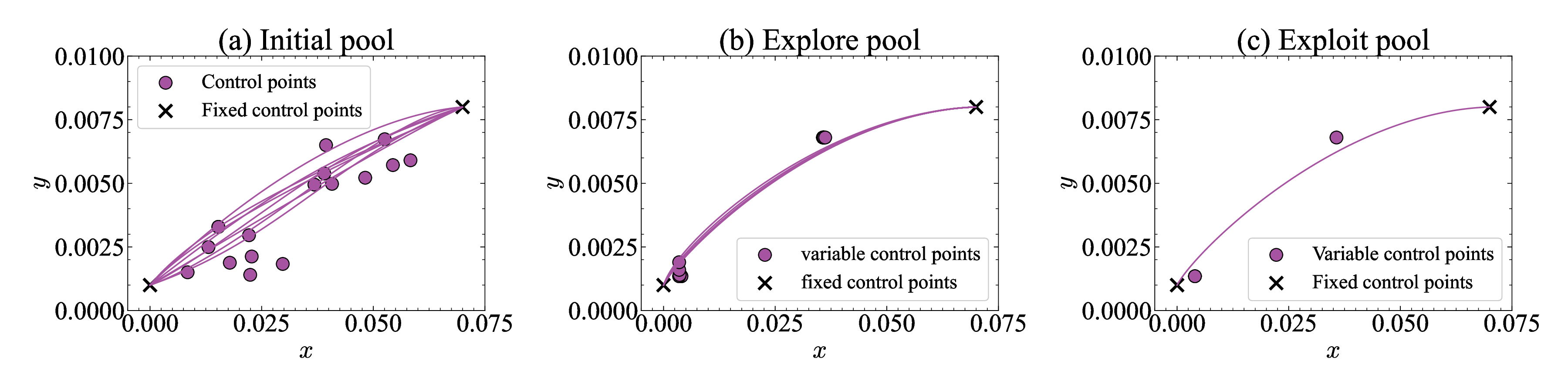}
        \caption{(a) Initial pool of solutions (b) explore pool of solution, and (c) exploit pool of solutions obtained from LEO}
        \label{nozzle_shapes_LEO}
\end{figure}

Figure \ref{nozzle_shapes_LEO}(a) shows the juxtaposition of the initial nozzle shapes that are fed to both explore and exploit pools. It is clear that the initial pool of solutions portrays significant variations among themselves without any bias towards any particular shapes, which is key for any optimisation task. Figure \ref{nozzle_shapes_LEO}(b) presents the final nozzle shapes from the explore pool at the end of 30 optimisation iterations, which exhibit minor variations, as expected. The optimal shape obtained from the exploit pool with the least radial velocity at the nozzle exit is shown in Fig. \ref{nozzle_shapes_LEO}(c). It is noted that the nozzle shape obtained from the final explore and exploit pool clearly demonstrated \textit{bell} shaped contours, which augers well with the expected shapes derived using isentropic flow theory \cite{rao1958exhaust}. This demonstrates that LEO is able to refine its search towards obtaining optimal shapes that are consistent with expected theory, albeit for a 4D problem. 



\begin{figure}
        \centering
        \includegraphics[width=0.75\textwidth]{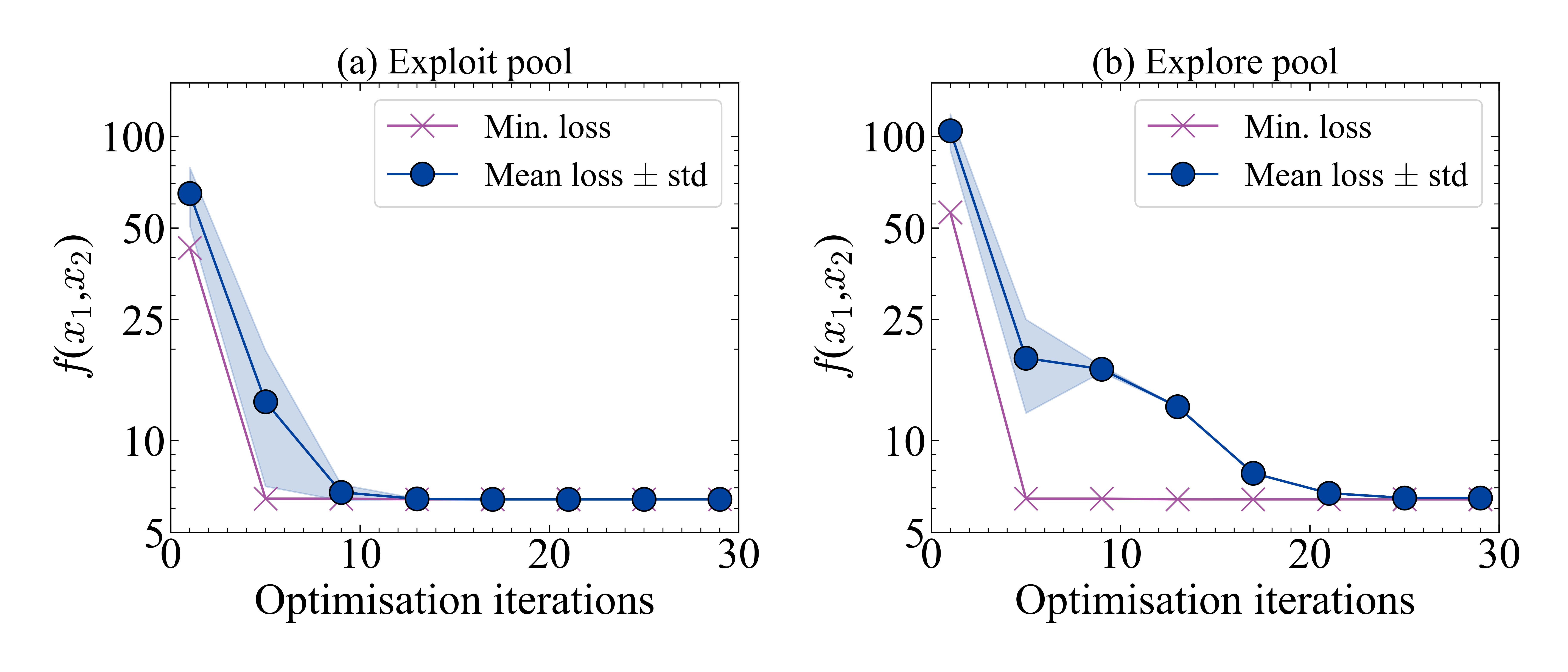}
        \caption{Convergence of the minimum and mean function values with optimisation iterations for exploit and explore pools}
        \label{nozzle4}
    \hfill
\end{figure}

\subsubsection{One-dimensional heat transfer problem}

Heat transfer is a fundamental phenomenon in many industrial applications. While robust numerical methods like finite difference coupled with iterative approaches allow us to compute the discretized solution to the governing equations \ref{heat_transfer_theory}, the same can be viewed as an optimisation problem, \textit{i.e.}, temperature distribution that results in minimization of the residual loss of the underlying governing equation \ref{1dheateqLoss}s. In this section, we determine the solution to a 1D steady-state heat transfer problem in a domain discretised using $N$ points using LEO. All the test cases in this section are solved using a population size of 10 for a total of 100 optimisation iterations. 

\begin{figure}
    \centering
    \includegraphics[width=\textwidth, height=\textheight,keepaspectratio]{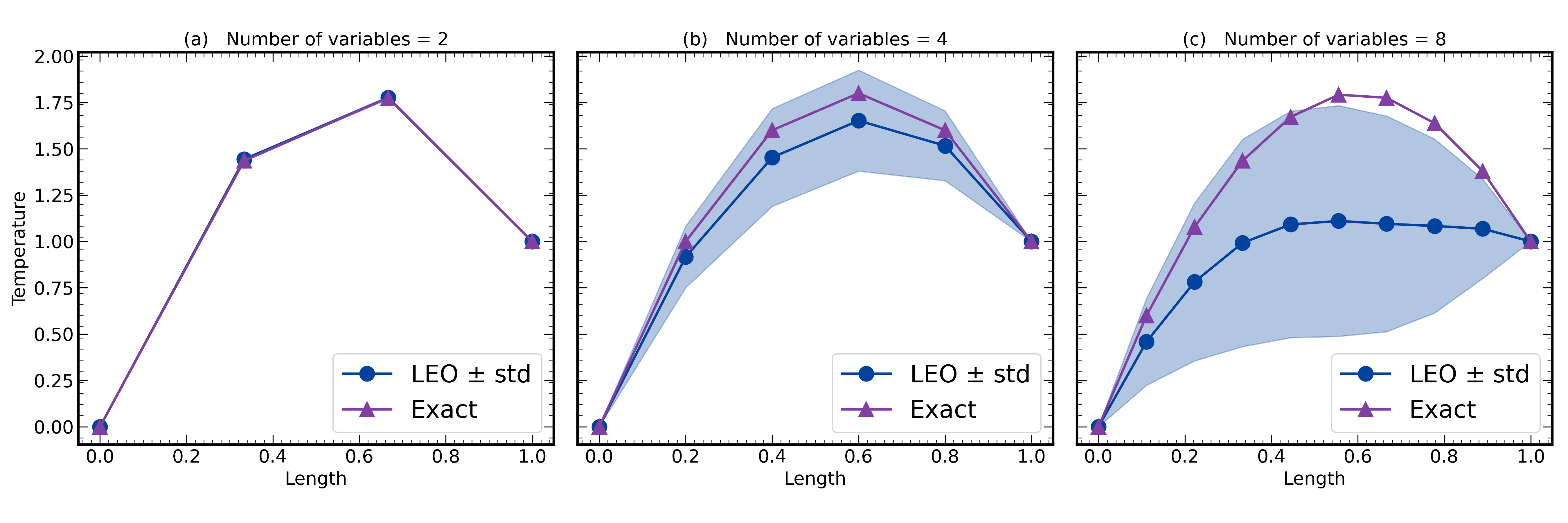}
    \caption{Heat equation optimization for increasing number of variables. }
    \label{Fig:1DHeatEquationPlots}
\end{figure}

Figure \ref{Fig:1DHeatEquationPlots} (a) shows the steady-state temperature distribution obtained by LEO and its comparison with the exact solution for $N=2$. While $N=2$ results in a rather jagged temperature profile, it is clear that the solution obtained from LEO agrees quite well with that of the exact solution. This signifies that a solution for a low-dimensional problem can result in a physics-consistent solution. Figure \ref{Fig:1DHeatEquationPlots} (b) shows the steady-state temperature profile for $N=4$ where minor observable differences with the exact temperature profile are noted. This is not surprising and rather points to the observation that has been made in Section \ref{high_dim_results}. Similarly, Fig. \ref{Fig:1DHeatEquationPlots} (c) shows a stark contrast between solutions obtained from LEO and their comparison with the exact temperature profile. This clearly indicates that while the present approach performs remarkably well in yielding physics-consistent solutions for low-dimensional problems, the nature of the solution starts to deviate significantly for higher degrees of freedom, necessitating the need to evolve the search with higher number of optimisation iterations using higher population size. 

\subsubsection{Windfarm layout optimization}
In this experiment, we will investigate the capabilities of LEO in optimizing the layout of a wind farm, which is essentially the process of finding the optimal locations of wind turbines in a wind farm to maximize the power production \ref{windfarm_theory}. \\

\begin{figure}
    \centering
    \includegraphics[width=\textwidth, height=\textheight,keepaspectratio]{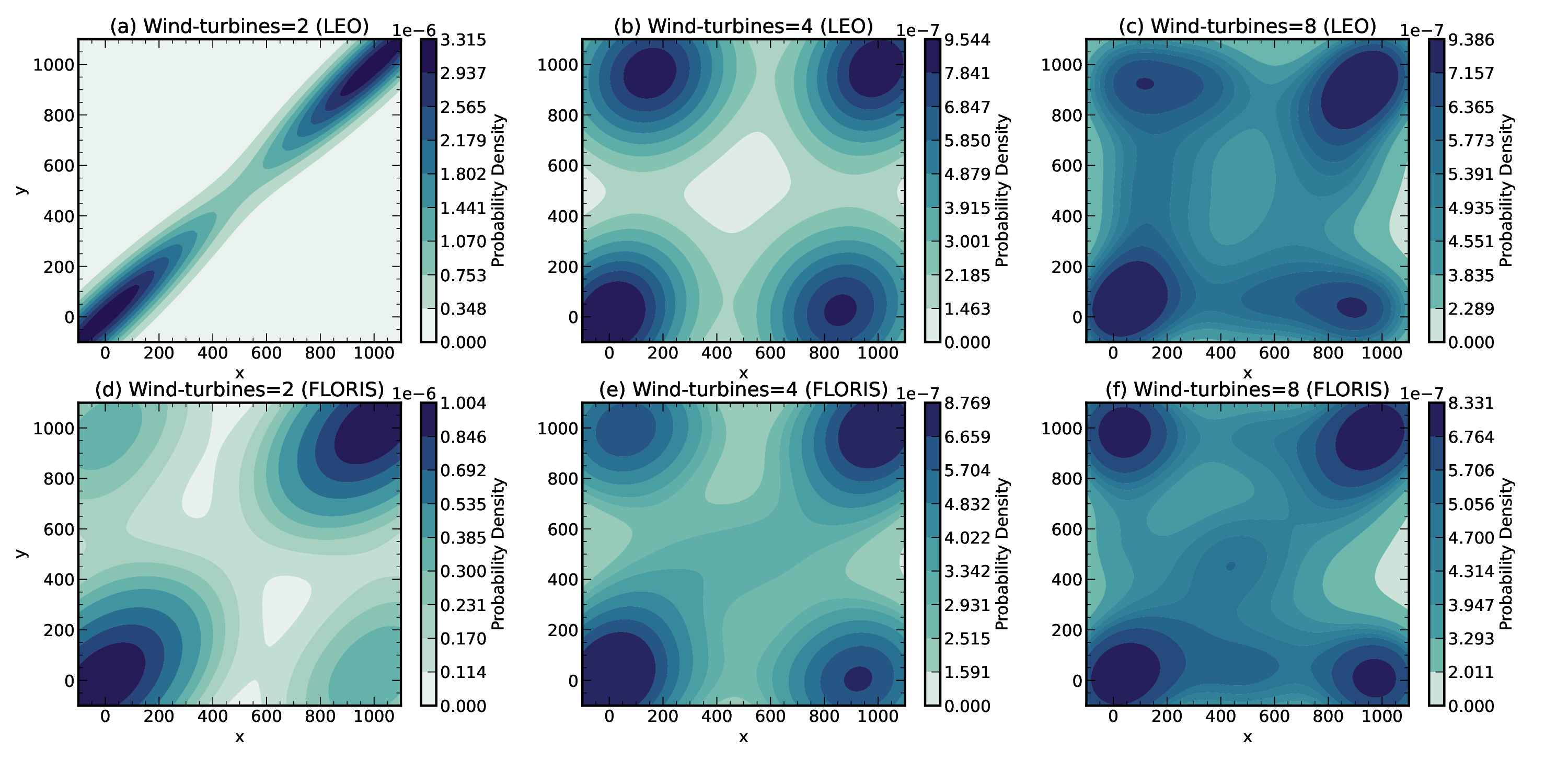}
    \caption{Windfarm layout optimization for increasing number of wind-turbines. }
    \label{Fig:WindfarmLayoutOptPlots}
\end{figure}

\begin{figure}
    \centering
    \includegraphics[width=\textwidth, height=\textheight,keepaspectratio]{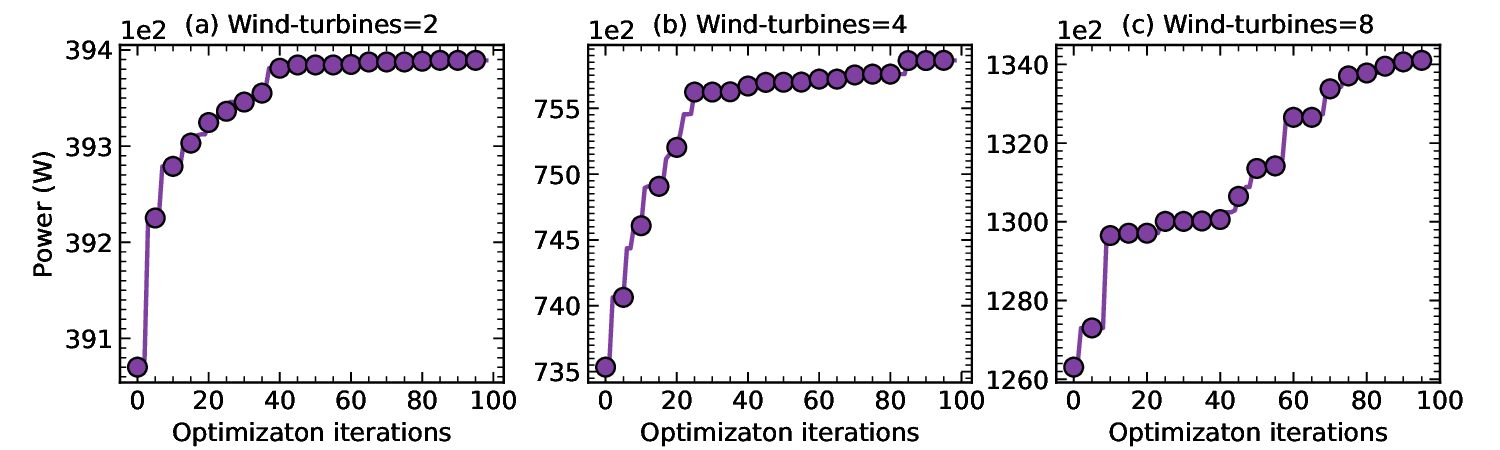}
    \caption{Windfarm layout optimization loss curves for increasing dimensionality of the problem.}
    \label{Fig:WindfarmLayoutOptLossPlots}
\end{figure}

Figure~\ref{Fig:WindfarmLayoutOptPlots} shows the kernel density estimation (KDE) obtained for the optimal layout considering all 30 runs in each of the three cases with 2, 4, and 8 turbines within the domain described previously. This would result in 4, 8, and 16 decision variables, respectively, for the optimization problem as each turbine is uniquely specified in a 2D domain with $(x, y)$ coordinates. In the layout optimization routine of FLORIS, Scipy's SLSQP optimization method is used as the solver. We can clearly see from this figure that LEO is able to produce layouts with a distribution that is in agreement with what is obtained with conventional Scipy's SLSQP method.
Figure~\ref{Fig:WindfarmLayoutOptLossPlots} shows the Annual Energy Production (AEP) value with iterations for a randomly picked run in each of the three cases. This figure clearly demonstrates the optimization trajectory as the iterations progress, demonstrating that LEO can indeed optimize industrial scale problems.

%% file: 4_discussion.tex
In this section, we explore the reasoning capabilities of LLMs backed by strong evidence. We show that LLMs inherently possess attributes of reasoning that, when assisted by an \textit{elitism} criterion, allow for a hybrid framework that renders faster convergence towards a global optimal solution for highly non-convex optimisation problems. We then list out various challenges that are encountered while using LLMs as optimizers. This is followed by enumerating various engineering tweaks and strategies that were devised in conjunction with LLMs that allow us to circumvent such issues. In our experience, these tweaks have been instrumental in getting the best optimization performance out of the LLMs. Finally, we provide some future directions towards improving the performance of LEO for high dimensional problems.

\subsection{Proof of LLM reasoning capabilities}
The elitism approach that we adopt in this paper, i.e., the explore-exploit-port-filter strategy, naturally prompts the question, ``Does LLM's reasoning indeed result in better candidate solutions, resulting in faster convergence to global optimal solution, or would a LLM tasked with the generation of random exploit and explore solutions result in better convergence?" 
To systematically address this issue, we designed the following two experiments: (a) In the first experiment, we examine the performance of LLM assisted by guardrails (\textit{i.e.,} LEO) and compare with the same framework where LLM are used to randomly generate explore and exploit solutions without guardrails (LEO-Rnd) (b) In the second experiment, we undertake additional statistical evaluations to exhibit the reasoning component of the LLM in arriving at the final optimal solution. We discuss our findings in the next two sub-sections. 

\subsubsection{Experiment 1: LLM solely as solution generator}
In this experiment, we designate the LLM as the role of only a random solution generator (LEO-Rnd), \textit{i.e.}, random perturbation of candidate points, without any historical context or guardrails such as elitism (refer Tables \ref{explore_prompt} and \ref{exploit_prompt}). The context in question is the information about candidate solutions and their function values from previous optimisation iterations. In doing so, the burden of finding the optimal solution falls completely on the $\texttt{port\_and\_filter}$ function. This exercise also allows us to prove the contrary: LLMs, when provided with the context of candidate solutions and tasked with the generation of new candidate solutions such that function value is reduced, result in faster convergence (refer to Table \ref{Table:b_vs_c_table}). The purpose of this experiment is to explicitly demonstrate the superiority of LLMs to generate better candidate points using optimization history (LEO), as opposed to simply using LLM to exhaustively generate solutions in an explore and exploit pool that might locate the global optimal solution (LEO-Rnd).

\begin{table}[]
\small
\begin{tabular}{llllll}
\hline
         & \textbf{Task}                                                                                                                                & \textbf{Context}                                                                                                    & \textbf{Guardrails} & \textbf{Explore Prompt}                                     & \textbf{Exploit Prompt}                                    \\ \hline
\textbf{LEO}      & \begin{tabular}[c]{@{}l@{}}Generate Solutions\\ at $k^\mathrm{th}$ iteration \\ such that function \\ value is reduced\end{tabular} & \begin{tabular}[c]{@{}l@{}}Solutions and \\function values\\ at $k$ - 1$^\mathrm{th}$ iteration\end{tabular} & Elitism    & Refer Table \ref{explore_prompt} & Refer Table \ref{exploit_prompt} \\ \hline
\textbf{LEO-Rnd} & \begin{tabular}[c]{@{}l@{}}Generate solutions at \\ $k^\mathrm{th}$ iteration\end{tabular}                                          & None                                                                                                       & None       & Refer Table \ref{explore_prompt} & Refer Table \ref{exploit_prompt} \\ \hline
\end{tabular}
\caption{\label{Table:b_vs_c} Difference between the underlying approach adopted for LEO and LEO-Rnd}
\end{table}

We show experiments using both LEO and LEO-Rnd in optimizing the 2D Goldstein-Price function. This test was performed for 100 optimisation iterations using 10 different random number seeds. Table \ref{Table:b_vs_c} highlights the differences between the two methods. Table \ref{Table:b_vs_c_table} shows the results obtained from this experiment, where it is found that LEO outperforms LEO-Rnd in terms of mean and median convergence with optimisation iterations. It must be noted that LEO is able to accurately identify the global optimal solution, whereas this is not reflected in solutions obtained from LEO-Rnd. In addition, Figs. \ref{Fig:ReasoningPlots}(a) and \ref{Fig:ReasoningPlots}(b) show the convergence plots of LEO and LEO-Rnd for both the mean and the median, computed across $10$ experiments. 
It can be observed that at the start of the optimisation, the rate of convergence exhibited by LEO is much higher than that of LEO-Rnd, which shows that it very quickly navigates from the poor initial guess to the global optima. On the contrary, LEO-Rnd shows signs of plateau, signifying that the LLM-based random solution generation without context and guardrails fails to locate the global optima, despite the $\texttt{port\_and\_filter}$ function serving as a 'soft' optimisation filter. We argue that this experiment is a clear demonstration of the ability of LLMs to build a topological relation of the decision variable with function space, with which it adaptively generates better candidate points, that aid in much faster convergence, thereby exhibiting reasoning capabilities.


\begin{table}[]
\begin{center}
\begin{tabular}{lll}
\hline
       & Mean $\pm$ std  & Median \\ \hline
LEO & 3.008 $\pm$ 0.019 & 3.000   \\ \hline
LEO-Rnd & 5.275 $\pm$ 1.718 & 5.407   \\ \hline
\end{tabular}
\end{center}
\caption{\label{Table:b_vs_c_table} Comparison of numerical accuracy of LEO and LEO-Rnd methods for $2D$ Goldstein Price problem}
\end{table}


\subsubsection{Experiment 2: Diminishing variance of the explore pool}

A question of paramount importance in this study is to clearly demonstrate the reasoning ability of LLM to yield better candidate solutions. An affirmative answer to such an evaluation will form the backbone of the present optimisation approach. To this extent, we ask the following question: if, in the present strategy, LEO progressively shows signs of diminishing variance in the explore pool, does it amount to reasoning? This is not a trivial proposition, as solutions are always exported from the explore pool to the exploit pool. So naturally, it is expected to have diminishing variance in the exploit pool and not the other way around. Additionally, if the explore pool progressively aligns in the direction of the global optimal solution, it naturally results in the generation of better candidate solutions in the exploit pool due to the use of the \texttt{port\_and\_filter} function. While several papers, for example \citep{liu2023large,yang2023large,guo2023optimizing}, use the temperature to control the explore and exploit behavior, we entrust the responsibility of generating explore and exploit pools to the model itself.


\begin{figure}
    \centering
    \includegraphics[width=1.1\textwidth, trim={2cm 4cm 0cm 0cm}, clip]{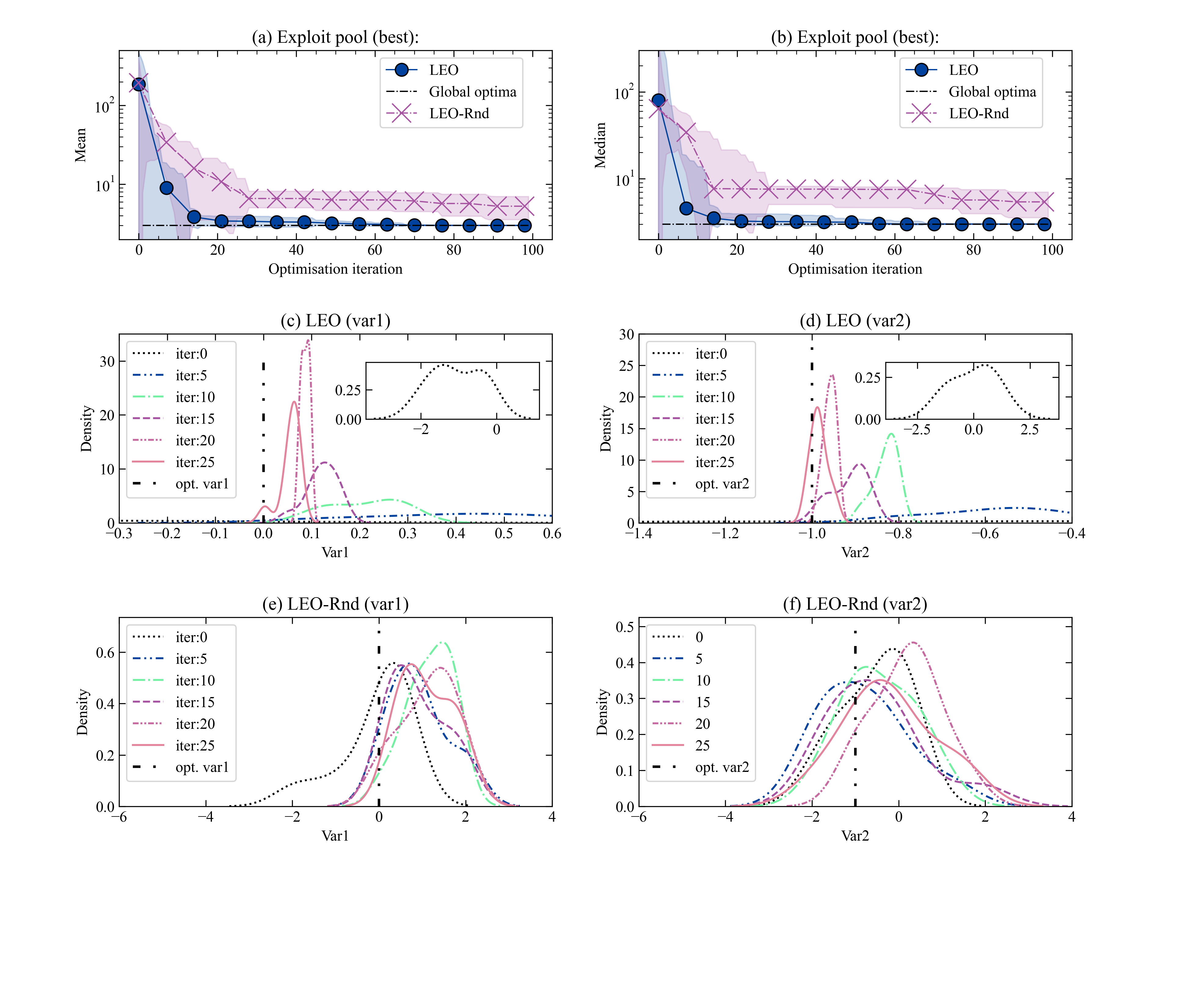}
    \caption{Plots demonstrating reasoning by LEO. Comparison of convergence behaviour depicted by LEO and LEO-Rnd in terms of (a) mean and (b) median with optimisation iterations. Comparison of kernel density functions for solutions belonging to explore pool (every 5th iteration) obtained from (c) LEO (var1), (d) LEO (var1), (e) LEO-Rnd (var1), and (f) LEO-Rnd (var2).}
    \label{Fig:ReasoningPlots} 
\end{figure}

Figure \ref{Fig:ReasoningPlots} presents the results for the Goldstein- Price problem. Specifically, Figures \ref{Fig:ReasoningPlots}(c) and \ref{Fig:ReasoningPlots}(d) illustrate the variation of the Kernel Density Function (KDF) for the explore pools corresponding to the two variables in the LEO method. Similarly, Figures \ref{Fig:ReasoningPlots}(e) and \ref{Fig:ReasoningPlots}(f) depict the KDF for the same two variables, but this time using the LEO-Rnd method.
Upon analyzing these figures, a clear trend emerges: the mean of the distribution associated with the explore pool in the LEO method converges towards the solutions, accompanied by a diminishing variance. This observation confirms that the model effectively reasons for and extracts `better' points from regions characterized by lower objective function values.
In contrast, the LEO-Rnd method does not exhibit a similar trend. This discrepancy can be attributed to the random selection of the explore pool. Notably, this finding underscores an important point: relying solely on temperature as a means to control exploration may not yield optimal results, as it primarily introduces randomness. On the other hand, the elitism-based guardrails effectively harness the reasoning capability of LLMs to generate `better' points.

\subsection{Challenges involved with LLM based Optimizers}
We now discuss some practical challenges we have encountered during our exploration of LLMs in the context of optimization:

\begin{enumerate}

\item \textbf{Reproducibility of exact performances}: We can fix a random seed to start the optimization from the same point. However, we cannot curtail the randomness. If we fix the temperature to zero, we can ensure repeatability, but at the cost of reduced performance due to poor exploration. Alternatively, a higher temperature value would improve the model's performance at the cost of reduced repeatability. Further, as argued previously, a very high temperature value can result in diminished performance of the method. 

\item \textbf{LLM hallucinations}: LLMs sometimes produce numbers such as $1.23456$ (i.e., some common sequences) or common integers such as $0$ or $1$, which are not backed by any rationale. This issue has been reported in \cite{hopkins2023can}.  
 This sometimes leads to the model getting stuck at local minima. 

\item \textbf{Mode collapse problem}: In spite of a higher temperature value, there are times at which a LLM keeps producing the same output, often a trivial number such as $1.0000$ or $0.0000$. This stems from the model-collapse issue pertaining to the auto-regressive nature of the algorithm, as explained in \cite{hopkins2023can}.

\item \textbf{The impact of increasing dimensions on model accuracy}: From benchmark problems to engineering applications, we consistently observe that LEO produces erroneous solutions for high-dimensional cases. Further, the uncertainty in the solutions increases with an increase in dimensionality.

\item \textbf{Computational expenses of large language models}: As this technology is still in its infancy, there are currently high costs associated with querying GPT-3.5 Turbo, which is used in this study. More advanced models, such as the GPT-4, are even more expensive. However, as the technology evolves towards maturity, it is anticipated that the costs associated with LLMs will decrease, concurrently improving their efficacy. 

\item \textbf{Limitations in addressing saddle points}: This method may not be able to identify and solve saddle point problems, because the explore-exploit strategy is based on prompts instructing the user to minimize the objective function value. 

\end{enumerate}

\subsection{Remedies}
Some of the remedies that would be helpful to mitigate the above mentioned challenges are as follows: 
\begin{enumerate}
\item Exact reproducibility is still unsolved, but a straightforward way to extract consistent performance is to follow an ensemble approach. The mean of several runs converges to the global optima as the number of samples increases.  
\item To address issues such as mode collapse and hallucinated numbers, we add `jitter' to the prompt. This means a small amount of perturbation added to the examples in the prompts seems to solve both of these issues. 
\item Assigning the identity of an optimization researcher at the very beginning of the prompt seems to improve the performance. 

\end{enumerate}
The detailed prompts used for the explore and exploit operations are presented in Appendix B.

\subsection{Improving accuracy for high-dimensional problems}
A possible remedy for the high-dimensional problem could be the stochastic sampling suggested in \citep{hu2023tackling}. Herein, a subset of dimensions, randomly chosen, are handled at a time by LEO. The batch-size becomes a hyperparameter and would depend on the context-window length of the LLM. Another possible approach to tackling the high-dimensional problems could be an agent based approach. Here, a number of LEO agents can work on a smaller number of dimensions and communicate their outputs through a message-passing interface or graph-based framework. A successful solution strategy for the high-dimensional problems could have a huge impact on the field of scientific machine learning, and would be the topic of future research.

%% file: 5_conclusion.tex
This paper presents a population-based optimization method based on LLMs called Language-model-based Evolutionary Optimizer (LEO). We present a diverse set of benchmark test cases, spanning from elementary examples to multi-objective and high-dimensional numerical optimization problems. Furthermore, we illustrate the practical application of this method to industrial optimization problems, including shape optimization, heat transfer, and windfarm layout optimization.

Several key conclusions can be drawn based on the obtained results. 
\begin{enumerate}
    \item We assert that LLMs exhibit the capacity to reason and execute zero-shot optimization. The explore-exploit strategy, as proposed in this paper, effectively harnesses this inherent ability and consistently performs on par with, if not surpassing, some of the leading methods in numerical optimization.
    \item We deduce that while LLMs are capable of optimization, achieving consistent and improved performance necessitates a series of engineering adjustments. LLMs tend to exhibit creativity and a propensity for hallucinations. Consequently, prompt engineering interventions become crucial to eliciting consistent behavior
from these models. 
\item Our observations reveal that although LLMs excel in low-dimensional problems, they require a greater number of iterations to converge to a global optimum in high-dimensional scenarios. 
This phenomenon is reflected in a gradual reduction in accuracy as the dimensionality increases for the same 
fixed number of iterations, a trend also observed in the engineering examples. 
\item LEO offers several advantages over conventional methods for low-dimensional problems, such as a parameter-free method, ease of implementation, suitability for multi-objective optimization problems with conflicting objectives, as well as problems with multiple minima, automatic step-size selection, etc. 

\item We have demonstrated that LLM-based exploration and exploitation offers a straightforward way to yield `effects' of crossover and mutation akin to evolutionary strategies such as the popular NSGA II approach. This enables a seamless way to perturb candidate solutions that distinctively retain the identity of the Pareto optimal front. 

\end{enumerate}

While the method still requires overcoming a few hurdles to make it suitable for engineering applications, such as tackling high-dimensional problems and the high costs associated with running LLMs, as discussed previously, it still offers the following advantages in its current form:
\begin{enumerate}
    \item \textbf{Parameter-Free Method:} Our approach operates without the need for any user-defined parameters. This inherent flexibility renders it suitable for a wide spectrum of problems, eliminating the burden of fine-tuning specific parameters. 
    \item \textbf{Decoupling Exploration and Exploitation:} Unlike other LLM-based optimization methods that rely on temperature-based exploration, our approach treats exploration and exploitation as distinct operations. This, when coupled with a rich context provided to the LLM, results in a superior distribution of candidate values. Notably, our experiments demonstrate that this approach leads to faster convergence compared to random point selection (governed by temperature) while maintaining other settings.
    \item \textbf{Unrestricted step size (learning rate):} Unlike other gradient-based optimizers, there is no restriction on step size (learning rate), since it is automatically inferred based on the samples presented in the prompt. This addresses issues associated with too small as well as too large learning rates, such as getting stuck in local minima and loss of stability.
    \item \textbf{Optimizing for multi-objective problems} It is suitable for multi-objective optimization problems, as well as for problems with multiple global minima, as evident from the test cases presented in this study. 
    \item \textbf{A modular approach:} Owing to its simplicity, it can be seamlessly incorporated into any existing population-based optimization framework. This vastly improves the usability of this method. 
\end{enumerate}
The scope of this paper is limited to exploring the present capabilities of LLMs, specifically GPT3.5 Turbo, in the context of
optimization. The effects of LLM-specific parameters such as model size, model architecture, and modality (single-modal vs. multi-modal) on the abilities of the models to perform numerical optimization will be explored in future work. 
Additionally, our forthcoming work will address strategies aimed at mitigating the curse of dimensionality.

%% file: 6_acknowledgement.tex
The authors from Shell India Markets Pvt. Ltd. would like to acknowledge the financial support from Shell. 

%% file: 7_appendix.tex
\subsection{Nose cone shape optimisation}
\label{nose_cone_theory}

\begin{figure}[htpb]
        \centering
        \includegraphics[width=0.5\textwidth]{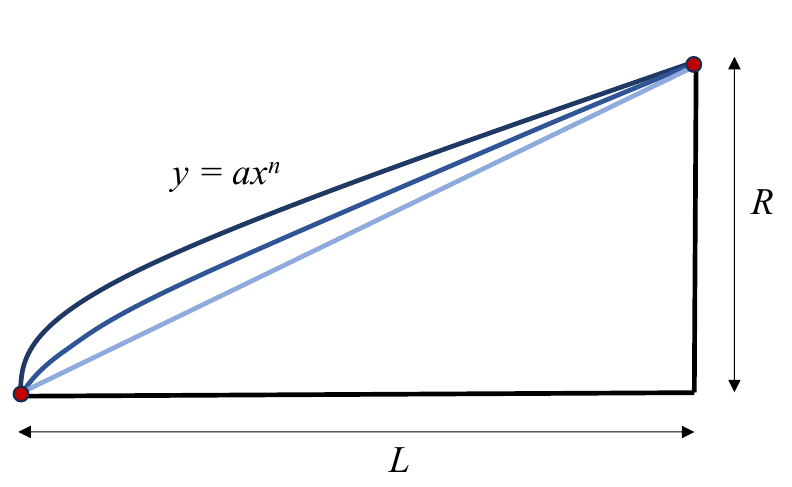}
        \caption{Juxtaposition of typical nose cones represented using a simple power law body}
        \label{nosecone_shape}
    \hfill
\end{figure}

The nose cone shape optimisation has been a well-studied problem in the aerospace industry since the early 1950's. Researchers have adopted various theoretical approaches, such as the Newtonian method, Taylor-Maccoll theory, the Fay-Riddel stagnation point heat transfer approach, etc., to design nose cone shapes with the objective of minimizing drag coefficients, ballistic coefficients, heat transfer, etc. \cite{brahmachary2018maximum}. In the present work, we represent the shape of the nose cone body having a fineness (or length to diameter) ratio of $L/D$=1 using a simple power-law curve as follows:
\begin{equation}\label{Eq:nosecone}
 y = ax^n \ ; \ 0.001 \leq n \leq 1
\end{equation}
where $n$ is the decision variable that determines the shape of the body (see Fig. \ref{nosecone_shape}). In addition, we employ Newtonian theory to compute the drag coefficient $C_d$. 


\subsection{Nozzle shape optimisation}
\label{nozzle_theory}

\begin{figure}
        \centering
        \includegraphics[width=0.75\textwidth]{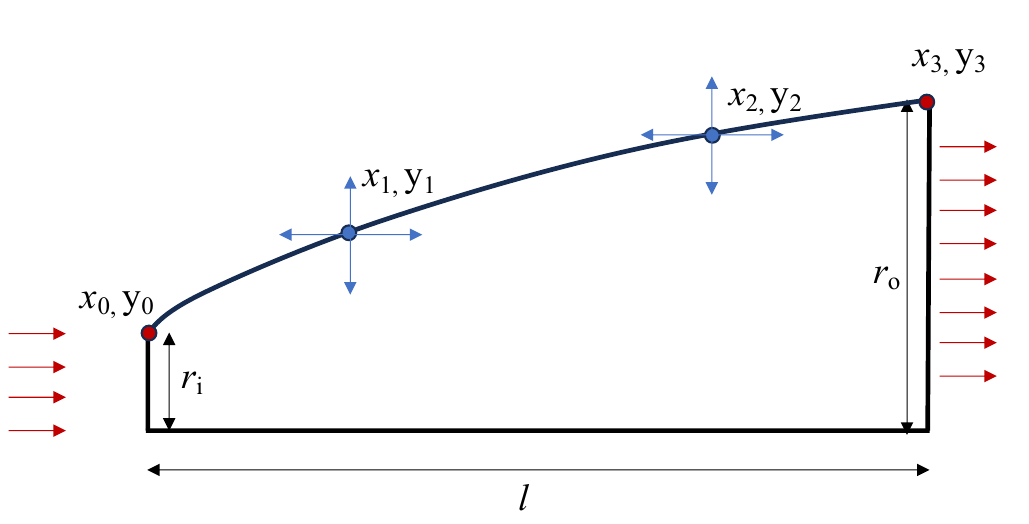}
        \caption{Schematic of the nozzle shape representation and associated variable}
        \label{nozzle_shape}
    \hfill
\end{figure}

The nozzle shape design is yet another key application in the aerospace industry. For instance, nozzle shape design is an integral component in the design of ramjet/scramjet engines for high thrust. In the present study, the two-dimensional (2D) nozzle is designed with the objective to maximise the streamwise velocity or minimise the radial velocity \cite{dudas2020high}. For any given nozzle shape, the radial velocity or the streamwise velocity can be easily computed using a simple algebraic equation such as the area-Mach number relation, invoking the simple quasi-1D flow assumptions for high-speed compressible flows. The 2D nozzle in consideration is parameterized using a cubic Bezier curve as follows:
\begin{align*}
x(t) &= (1-t)^3x_0 + 3(1-t)^2tx_1 + 3(1-t)t^2x_2 + t^3x_3, \\
y(t) &= (1-t)^3y_0 + 3(1-t)^2ty_1 + 3(1-t)t^2y_2 + t^3y_3, \quad t \in [0,1].
\end{align*}
where the control points ($x_0, y_0$) and ($x_3, y_3$) are fixed in space (for fixed inlet radius, $r_i$, outlet radii, $r_o$, and length, $l$), whereas the remaining control points ($x_1, y_1$) and ($x_2, y_2$) are variables (see Fig. \ref{nozzle_shape}). This results in a 4D problem. 

\subsection{One-dimensional heat transfer problem}
\label{heat_transfer_theory}

We attempt to solve a 1D steady-state heat transfer equation as shown below:
\begin{equation}
\frac{d^2 T}{dx^2} + \frac{q_x}{k} = 0 \ | \ T_L=0, \ T_R=1
\end{equation}
\label{1dheateq}
where $T$ is the non-dimensional temperature distribution and $q/k$ is the internal heat generation per unit volume. In order to solve the steady-state temperature distribution that satisfies the above equation, we impose the Dirichlet boundary condition at the extreme ends of the 1D computational domain, as shown above. The domain of length $L$ is discritised using $N$ elements, which are varied across multiple test cases in this present experiment. This results in an equal elemental spacing of $\Delta x$ = $L/N$. We adopt second order central differencing for the second order term, followed by a constant value for $q/k$ = 10. This results in a residual loss term (or objective function) as follows:\\
\begin{equation}
\frac{T_{i-1}-2T_i+T_{i+1}}{\Delta x^2} + \frac{q_x}{k} = 0
\end{equation}
\label{1dheateqLoss}

While there are established iterative methods that allow us to compute the steady-state temperature distribution across the 1D computational domain, we frame this problem as an optimisation task. This results in an approach wherein we determine the discretized temperature field that admit solution to Eq. \ref{1dheateq}, i.e., minimization of loss $\sim 0$.

\subsection{Wind farm layout optimisation}
\label{windfarm_theory}
Wind farm layout optimisation is a critical endeavor in the renewable energy sector. This optimisation task is aimed at enhancing the efficiency and output of the wind farms by manipulating the position of the wind turbines. The objective function in this case is the annual energy production (AEP), which is the amount of electricity that a wind farm can generate in one year. AEP is influenced by wind conditions such as wind direction and speed. We consider 72 wind directions ranging from $0^{\circ}$ to $360^{\circ}$ with a step size of $5^{\circ}$, and we consider a random wind speed whose value is centered around $8~m/s$ with a standard deviation of $0.5~m/s$. The rotor diameter of the turbine is taken to be $126.0~m$ and the domain is considered to be of size $1000~m \times 1000~m$. The wake calculation settings are kept the same across all the runs.

%% file: 8_promptappendix.tex
\begin{table}[htpb]
\scriptsize
\centering
\begin{tabular}{|l|l|l|}
\hline
                                                                   & \textbf{LEO}        & \textbf{LEO-Rnd}           \\ \hline
\textbf{\begin{tabular}[c]{@{}l@{}}Explore \\ Prompt\end{tabular}} & \begin{tabular}[c]{@{}l@{}}You are an optimization researcher \\ tasked to minimize the value of loss. \\ Current candidate solutions for \\ N=2 variables in the order \\ var1, var2 with their respective \\ function loss in csv format are:\\ \\ var1, var2, loss\\ \\ -0.018822,-0.692810,74.271916\\ 0.320219,-1.629214,3430.938028\\ 0.886970,-1.846776,3507.116813\\ -1.149739,0.407518,4330.723936\\ 1.958779,-0.219516,5479.393434\\ -0.489178,0.695195,22743.401171\\ -1.957878,0.271882,91001.711544\\ -0.095039,1.589113,130107.754358\\ -0.664592,1.331668,169337.055368\\ -1.622905,1.526054,353181.485985\\ \\ You have to look at the above points \\ and think of the (min, max) values \\ for each variable that might reduce \\ or minimize the loss. With these limits \\ of (min, max) values for each variable, \\ you must provide exactly 10 new, but \\ completely different and scattered sets \\ of values from the above ones, to \\ explore away regions for minimizing \\ the function loss. Generate the result \\ like a csv file with 10 rows and 2 \\ columns, where each row represents \\ a candidate and each column represents \\ a variable. The response must only \\ contain these numerical values in the \\ csv format without column headers. Do \\ not provide additional text or explanation. \\ Strictly, provide only the variable values \\ as floating point numbers with precision \\ format \%.6f.\end{tabular} & \begin{tabular}[c]{@{}l@{}}You must provide exactly 10 new, but \\ completely different and scattered sets \\ of values within all variables, in between \\ -2 and 2. Generate the result like a csv \\ file with 10 rows and 2 columns, where each \\ row represents a candidate and each column \\ represents a variable. The response must \\ only contain these numerical values in the \\ csv format without column headers. Do not \\ provide additional text or explanation. Strictly, \\ provide only the variable values as floating \\ point numbers with precision format \%.6f.\\ \\ For example (with column headers):\\ \\ 0.8166, 0.1868\\ 0.6788, 0.0772\\     ...\\ 0.1853, 0.3935\end{tabular} \\ \hline
\end{tabular}
\caption{Explore prompt for LEO and LEO-Rnd methods for Goldstein Price problem}
\label{explore_prompt}
\end{table}
\begin{table}[htpb]
\scriptsize
\centering
\begin{tabular}{|c|l|l|}
\hline
                                                                   & \multicolumn{1}{c|}{\textbf{LEO}}                 & \multicolumn{1}{c|}{\textbf{LEO-Rnd}}     \\ \hline
\textbf{\begin{tabular}[c]{@{}c@{}}Exploit \\ prompt\end{tabular}} & \begin{tabular}[c]{@{}l@{}}You are an optimization researcher \\ tasked to minimize the value of loss. \\ Current best candidate solution for \\ N=2 variables in the order \\ var1, var2, with their respective \\ function loss in csv format are:\\ \\ var1, var2, loss\\ \\ 1.8859, 0.2995, 101.5838\\ \\ Please provide exactly 10 new but different \\ candidates of values for all variables, in \\between -2 and 2, to exploit close by regions \\for minimizing the function loss. Generate \\the result like a csv file with 10 rows and 2 \\columns, where each row represents a candidate \\and each column represents a variable. The \\response must only contain these numerical \\values in the csv format without column \\ headers and row index. Do not provide \\additional text or explanation. Strictly, \\ provide only the variable values as floating \\point numbers with precision format \%.6f.\\ \\ For example (with column headers):\\ 0.8166, 0.1868\\ 0.6788, 0.0772\\     ...\\ 0.1853, 0.3935\end{tabular} & \begin{tabular}[c]{@{}l@{}}Please provide exactly 10 new but \\ different candidates of values for all \\ variables, in between -2 and 2 to \\ exploit close by regions. Generate \\ the result like a csv file with 10 rows \\ and 2 columns, where each row \\ represents a candidate and each \\ column represents a variable. The \\ response must only contain these \\ numerical values in the csv format \\ without column headers and row \\ index. Do not provide additional \\ text or explanation. Strictly, provide \\ only the variable values as floating \\ point numbers with precision format \%.6f.\\ \\ For example (with column headers):\\ \\ 0.8166, 0.1868\\ 0.6788, 0.0772\\     ...\\ 0.1853, 0.3935\end{tabular} \\ \hline
\end{tabular}
\caption{Exploit prompt for LEO and LEO-Rnd methods for Goldstein Price problem}
\label{exploit_prompt} 
\end{table}
\begin{table}[htpb]
\scriptsize
\centering
\begin{tabular}{|l|l|}
\hline
                                                                   & \textbf{LEO}         \\ \hline
\textbf{\begin{tabular}[c]{@{}l@{}}Exploit \\ Prompt\end{tabular}} & \begin{tabular}[c]{@{}l@{}}You are an intelligent assistant who can understand great technical details. \\ You will help me minimise two functions by taking cues from given information. \\ Current candidate solutions (a, b) with their function loss are:\\ \\ Candidate 1: a=0.2536, b=0.7346, function loss f1=0.2536, function loss f2=6.2219\\ Candidate 2: a=0.3953, b=0.1976, function loss f1=0.3953, function loss f2=1.7307\\ \\ Please provide exactly 2 new but different pairs of values for 0\textless{}'a'\textless{}1 and 0\textless{}'b'\textless{}1 \\ to exploit closeby regions for minimizing the function loss1 and function loss2. Take \\ cues from the above points. The response should contain 4 values, corresponding \\ to 2 pairs in the format: 'a1, b1, a2, b2, ..., an, bn'. Provide only these 4 numerical \\ values in order as python {[}list{]}, separated by commas, with no additional text or \\ explanation.\end{tabular} \\ \hline
\textbf{\begin{tabular}[c]{@{}l@{}}Explore \\ Prompt\end{tabular}} & \begin{tabular}[c]{@{}l@{}}You are an intelligent assistant who can understand great technical details. \\ You will help me minimise two functions by taking cues from given information. \\ Current candidate solutions (a, b) with their function loss are:\\ \\ Candidate 1: a=0.3953, b=0.1976, function Loss f1=0.3953, function Loss f2 =1.7307\\ Candidate 2: a=0.1061, b=0.9373, function Loss f1=0.1061, function Loss f2 =8.4352\\ \\ Please provide exactly 2 new but different pairs of values for 0\textless{}'a'\textless{}1 and 0\textless{}'b'\textless{}1 \\ to explore away regions for minimizing the function loss1 and function loss2. The response \\ should contain 4 values, corresponding to 2 pairs in the format: 'a1, b1, a2, b2, ..., an, bn'. \\ Provide only these 4 numerical values in order, separated by commas, with no additional \\text or explanation.\end{tabular}                                                        \\ \hline
\end{tabular}
\caption{Exploit and exploit prompts for NSGA II with LLM plug and play.}
\label{prompt_NSGA}
\end{table}

\begin{table}[htpb]
\scriptsize
\centering
\begin{tabular}{|p{0.2\linewidth}|p{0.8\linewidth}|}
\hline
                & \textbf{Prompt}                                                             \\ \hline
\textbf{Nosecone design} & \begin{tabular}[c]{@{}l@{}} You are an optimization researcher tasked to minimize the value of function \\loss. Current best candidate solution for 1 variable (n) with their respective \\ function loss are:  \\\\ candidate: n=0.90000; loss: 0.37450\\ \\ Give me a new (n) value that satisfies the following: (a) new n is different \\ from all above, (b) has a function value lower than the above function  \\ values and (c) result in a rapid convergence towards the value of n that \\ results in lowest function value. Do not write code or any explanation. The \\ output must end with just a numerical value for next variable.\end{tabular}                                                                      \\ \hline
\textbf{Rosenbrock 2D}   & \begin{tabular}[c]{@{}l@{}}You are an optimization researcher tasked to minimize the value of function \\loss with two input variables n1 and n2. Current best candidate solution for \\ 2 variables (n1, n2) with their respective function loss are:\\ \\ input: n1=1.00, n2=0.300, loss:49.00\\ \\ Give me a new (n1, n2) pair value that satisfies the following: (a) new n1 and \\ n2 is different from all above, (b) has a function value lower than the above \\ function values and (c) result in a rapid convergence towards the value of n1 \\ and n2 that results in lowest function value. Do not write code or any \\ explanation. The output must end with two numerical values for (n1, n2) only.\end{tabular} \\ \hline
\end{tabular}
\caption{Prompts for testing reasoning abilities of LEO for Rosenbrock $2D$ function and rocket nose cone shape optimization problem}
\label{prompt_reasoning}
\end{table}